\documentclass[10pt,twocolumn,letterpaper]{article}

\usepackage{cvpr}              %

\usepackage[dvipsnames]{xcolor}

\usepackage{graphicx}
\usepackage{booktabs}  
\usepackage{diagbox}
\usepackage{url}
\usepackage{multirow}
\usepackage{subcaption}

\usepackage{xparse}
\usepackage{xspace}
\usepackage{tikz}
\usepackage{multirow}
\usepackage{colortbl}
\usepackage{array}
\newcommand{\PreserveBackslash}[1]{\let\temp=\\#1\let\\=\temp}
\newcolumntype{C}[1]{>{\PreserveBackslash\centering}p{#1}}
\newcolumntype{R}[1]{>{\PreserveBackslash\raggedleft}p{#1}}
\newcolumntype{L}[1]{>{\PreserveBackslash\raggedright}p{#1}}

\usepackage{microtype}
\usepackage{graphicx}
\usepackage{mathtools}
\usepackage{subcaption}
\usepackage{booktabs} %

\usepackage{amssymb}%
\usepackage{pifont}%

\usepackage{bbold}

\usepackage{amssymb,enumitem}

\setlist[itemize]{leftmargin=*}
\setlist[enumerate]{leftmargin=*}

\renewcommand{\P}[1]{\operatorname{\mathbb{P}}[#1]}
\newcommand*{\rej}{{\ooalign{\lower.3ex\hbox{$\sqcup$}\cr\raise.4ex\hbox{$\sqcap$}}}}

\usepackage{stackengine}

\usepackage{xcolor}

\renewcommand{\ie}{\textit{i.e.,}\@\xspace}
\renewcommand{\eg}{\textit{e.g.,}\@\xspace}

\newcommand{\DMs}{DMs\@\xspace}
\newcommand{\DM}{DM\@\xspace}

\newcommand{\ours}{\texttt{CDI}\xspace} %
\newcommand{\ourslong}{\texttt{Copyrighted Data Identification}\xspace}

\renewcommand{\P}{\textbf{P}\xspace}
\newcommand{\U}{\textbf{U}\xspace}

\newcommand{\Pctrl}{\textbf{P}\textsubscript{ctrl}\xspace}
\newcommand{\Ptest}{\textbf{P}\textsubscript{test}\xspace}
\newcommand{\Uctrl}{\textbf{U}\textsubscript{ctrl}\xspace}
\newcommand{\Utest}{\textbf{U}\textsubscript{test}\xspace}

\newcommand{\Qsus}{\textbf{Q}\textsubscript{test}\xspace}

\DeclareRobustCommand\encircle[1]{\tikz[baseline=(char.base)]{\node[shape=circle,fill,inner sep=1pt] (char) {\textcolor{white}{#1}}}}

\graphicspath{{images/}}

\newcommand{\scoring}{scoring model\@\xspace}

\DeclareRobustCommand\encircle[1]{\tikz[baseline=(char.base)]{\node[shape=circle,fill,inner sep=1pt] (char) {\textcolor{white}{#1}}}}

\usepackage{booktabs,arydshln}
\makeatletter
\def\adl@drawiv#1#2#3{%
        \hskip.5\tabcolsep
        \xleaders#3{#2.5\@tempdimb #1{1}#2.5\@tempdimb}%
                #2\z@ plus1fil minus1fil\relax
        \hskip.5\tabcolsep}
\newcommand{\cdashlinelr}[1]{%
  \noalign{\vskip\aboverulesep
           \global\let\@dashdrawstore\adl@draw
           \global\let\adl@draw\adl@drawiv}
  \cdashline{#1}
  \noalign{\global\let\adl@draw\@dashdrawstore
           \vskip\belowrulesep}}
\makeatother

\newcommand{\nlp}[1]{}

\newcolumntype{x}[1]{>{\centering\arraybackslash\hspace{0pt}}p{#1}}

\usepackage{amsmath}

\DeclareMathOperator*{\argmin}{arg\,min}

\usepackage{amsmath,amsfonts,bm}

\def\eqref#1{equation~\ref{#1}}

\def\1{\bm{1}}

\def\eps{{\epsilon}}

\DeclareMathAlphabet{\mathsfit}{\encodingdefault}{\sfdefault}{m}{sl}
\SetMathAlphabet{\mathsfit}{bold}{\encodingdefault}{\sfdefault}{bx}{n}

\def\gD{{\mathcal{D}}}
\def\gE{{\mathcal{E}}}

\def\gR{{\mathbb{R}}}

\definecolor{cvprblue}{rgb}{0.21,0.49,0.74}
\usepackage[pagebackref,breaklinks,colorlinks,citecolor=cvprblue]{hyperref}

\def\paperID{10523} %
\def\confName{CVPR}
\def\confYear{2025}

\title{CDI: Copyrighted Data Identification in Diffusion Models}

\author{%
Jan Dubiński~\thanks{Equal contribution.}%
~~\thanks{Work done while the author was at CISPA.}\\
{\small Warsaw University of Technology, IDEAS NCBR}\\
{\tt\small jan.dubinski.dokt@pw.edu.pl}
\and
Antoni Kowalczuk~\footnotemark[1]\\
{\small CISPA Helmholtz Center for Information Security}\\
{\tt\small antoni.kowalczuk@cispa.de}
\and
Franziska Boenisch\\
{\small CISPA Helmholtz Center for Information Security}\\
{\tt\small boenisch@cispa.de}
\vspace{-0.15cm}
\and
Adam Dziedzic\\
{\small CISPA Helmholtz Center for Information Security}\\
{\tt\small adam.dziedzic@cispa.de}
\vspace{-0.15cm}
}

\begin{document}
\maketitle

\begin{abstract}
\vspace{-0.175cm}
Diffusion Models (DMs) benefit from large and diverse datasets for their training. Since this data is often scraped from the Internet without permission from the data owners, this raises concerns about copyright and intellectual property protections. While (illicit) use of data is easily detected for training samples perfectly re-created by a DM at inference time, it is much harder for data owners to verify if their data was used for training when the outputs from the suspect DM are not close replicas. Conceptually, membership inference attacks (MIAs), which detect if a given data point was used during training, present themselves as a suitable tool to address this challenge. 
However, we demonstrate that existing MIAs are not strong enough to reliably determine the membership of individual images in large, state-of-the-art DMs.
To overcome this limitation, we propose \ourslong (\ours), a framework for data owners to identify whether their \textit{dataset} was used to train a given DM. \ours relies on \textit{dataset inference} techniques, i.e., instead of using the membership signal from a single data point, \ours leverages the fact that most data owners, such as providers of stock photography, visual media companies, or even individual artists, own datasets with multiple publicly exposed data points which might all be included in the training of a given DM. By selectively aggregating signals from existing MIAs and using new handcrafted methods to extract features from these datasets, feeding them to a scoring model, and applying rigorous statistical testing, \ours allows data owners with as little as 70 data points to identify with a confidence of more than 99\% whether their data was used to train a given DM. Thereby, \ours represents a valuable tool for data owners to claim illegitimate use of their copyrighted data. 
We make our code available at {\url{https://github.com/sprintml/copyrighted_data_identification}}.
\vspace{-1cm}
\end{abstract}

\newpage
\section{Introduction}
\vspace{-0.175cm}

\begin{figure*}[t!]
    \centering
    \includegraphics[width=0.91\textwidth]{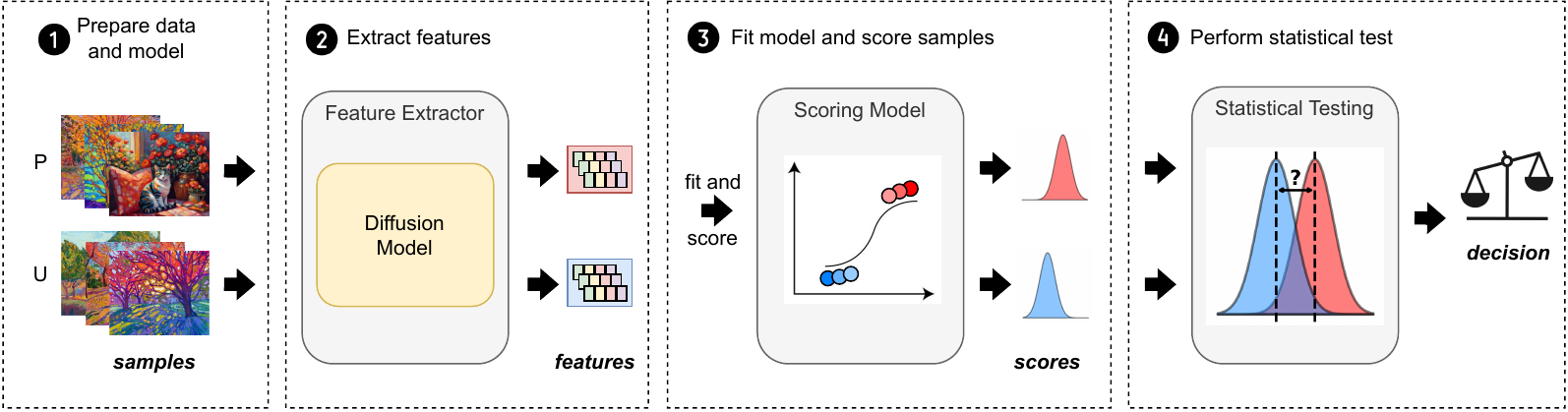}
    \caption{
    \textbf{Overview of our Dataset Inference method for Diffusion Models from our previous work.}
    Our approach consists of the following stages: \encircle{1}~Prepare the query data to verify if the \textit{published} suspect samples \P were used to train the \DM. The \textit{unpublished} samples \U from the same distribution as \P serve as the validation set. \encircle{2} Run inference on all the inputs $\{\P,\U\}$ to extract their membership features. Use current MIAs and our handcrafted features. \encircle{3} Find useful features and learn a discriminator. \P and \U sets are split into \Pctrl, \Ptest and \Uctrl, \Utest. The features for \Pctrl and \Uctrl are used to train a scoring model to selectively combine features and differentiate between the samples from \Ptest and \Utest.
    \encircle{4} Apply a statistical t-test to verify if the scores obtained for public suspect data point \P are statistically significantly higher than scores for \U, in which case, \P is marked as being a part of the \DM's training set. Otherwise, the test is inconclusive and the \DM's training set is resolved as independent of \P. The method
    }
    \label{fig:schema}
    \vspace{-0.5cm}
\end{figure*}

In recent years, large diffusion models (\DMs)~\cite{sohl2015deep} have rapidly
gained popularity as a new class of generative models, surpassing the performance of prior approaches, such as Generative Adversarial Networks~\cite{goodfellow2014generative}.
\DMs now power several state-of-the-art image generators including Stable Diffusion~\cite{rombach2022high}, Midjourney~\cite{midjourney}, Runway~\cite{rombach2022high}, Imagen~\cite{saharia2022photorealistic}, and DALL-E 2~\cite{ramesh2021zero, ramesh2022hierarchical}. 

To reach their powerful performance, \DMs need to be trained on large amounts of high-quality and diverse data.
This data is usually scraped from the Internet, often without respecting the copyrights of the data owners.
Especially since it has been shown that \DMs are capable of generating verbatim copies of their training data at inference time~\citep{carlini2023extracting}, this represents a violation of intellectual property rights.
Recently, Getty Images, a leading visual media company, filed a lawsuit against Stability AI, the creators of Stable Diffusion, alleging the unauthorized use of copyright-protected images~\citep{gety_v_stability,gettyimages_lawsuit}. This case has sparked a wave of additional lawsuits, with many more now addressing intellectual property infringement by generative AI companies~\citep{artists_lawsuit,artist_lawsuit}.
Unfortunately, as it becomes obvious during the lawsuits---particularly for training data points that are not output in a verbatim form during inference time---verifying that these data points have been illegitimately used for training the \DMs is a challenging task.

Membership inference attacks~\citep{DBLP:conf/sp/ShokriSSS17} that aim at identifying whether a specific data point was used to train a given model, in theory, present themselves as a solution to the problem. 
Unfortunately, prior work~\citep{dubinski2024towards} indicates that performing a realistic MIA on large \DMs is a very challenging task. 
One of the practical challenges lies in the prohibitive costs of training state-of-the-art \DMs (\eg \$600.000 for Stable Diffusion) which renders potent MIAs utilizing multiple \textit{shadow model} copies~\cite{DBLP:conf/sp/ShokriSSS17,carlini2022membership} infeasible. 
To further explore the practicality of MIAs for identifying copyrighted samples used to train large \DMs, we perform an extensive study, evaluating the success of existing MIAs against \DMs' training data for various open \DMs.
Our findings demonstrate that \textbf{current MIAs for \DMs are limited in confidently identifying \DMs' training data points in case of models trained on large datasets}---showcasing that individual MIAs cannot reliably support copyright claims.

In light of this result, we, however, observe that in most cases, data owners, such as stock photography, visual media companies, or even individual artists, typically seek to verify the use of not just a single data point but a collection of their work as training data for a given \DM.
This moves the idea of \textit{dataset inference}~\citep{maini2021dataset} (DI) into focus.
DI was first proposed to detect stolen copies of supervised classifier models and then subsequently extended to self-supervised models~\citep{dziedzic2022dataset}.
It leverages the observation that, while MIAs on individual data points do not produce a strong signal, selectively aggregating signals across a subset of the training dataset and applying statistical testing can reveal a distinct signature of the model. This dataset-based signature allows for the detection of stolen model copies with a confidence level exceeding 95\%.
Yet, to date, it remains unexplored whether the principles of DI actually transfer to \DMs and are suitable to identify subsets of their training data rather than resolving model ownership---given the vast amount of heterogeneous data \DMs are initially trained on.
Additionally, it is unclear how large the required training data subsets for verification would have to be.
Finally, we do not know the specific features needed to extract a strong signal over the training data.

To close these gaps, \textbf{we propose \ourslong, designed to answer the critical question: \textit{Was this \DM trained on a copyrighted collection of images?}} The overall schema of our method is illustrated in Fig.~\ref{fig:schema}. To design \ours, we move beyond simply aggregating features extracted by existing MIAs, as these often produce signals that are too weak to achieve highly confident DI, rendering the approach impractical. Instead, we firstly, extend the feature extraction methods by our newly proposed features. Secondly, we design a scoring function which maps the extracted information into sample membership probability, learning the features relevant for each \DM.  Finally, in contrast to MIAs which usually refer to metrics like True Positive Rate (TPR) or Area Under Curve (AUC) which do not give any confidence estimate, we equip \ours with rigorous statistical testing as the final component.

We demonstrate the success of our method on diverse large-scale \DM architectures (LDM~\citep{rombach2022high}, DiT~\citep{peebles2023scalable}, U-ViT~\citep{bao2023all}), including unconditioned, class-conditioned and text-conditioned models, trained on various image resolutions. 
Our results show that \ours achieves a confident detection rate of data (illegitimately) used for training \DMs.
\ours remains effective when only \textbf{a part of the investigated data was actually used in \DM training}.  
Moreover, we demonstrate that \ours does not yield false positives, making it a reliable tool for detecting and confidently claiming the use of copyrighted data in \DMs.

In summary, we make the following contributions:
\begin{itemize}
    \item We demonstrate that existing MIAs for \DMs show limited effectiveness in confidently identifying the training data points of large, state-of-the-art models.
    \item To address this issue, we propose \ours, a method that empowers data owners to identify whether their data has been (illegitimately) used to train a \DM,
    incorporating rigorous statistical testing to ensure confidence in the results.
    \item We perform thorough feature engineering to amplify the signal in \ours, proposing novel feature extraction methods and enabling data owners even with smaller datasets to benefit from our method.
    \item We evaluate \ours on a wide range of \DMs and their pre-training datasets and provide a unified open-source codebase\footnote{\url{https://github.com/sprintml/copyrighted_data_identification}} with a common interface to all prior MIAs and our new \ours, serving as a valuable evaluation testbed for the community.
\end{itemize}

\section{Background}

\textbf{Diffusion Models}~\cite{ho2020denoising,Song2022improvedScoreBasedGM} %
are generative models trained by progressively adding noise to the data and then learning to reverse this process. 
The forward diffusion process adds Gaussian noise $\epsilon \sim \mathcal{N}(0, I)$ to a clean image $x$ in order to obtain a noised image $x_t \gets \sqrt{\alpha_t} x + \sqrt{1 - \alpha_t} \epsilon$, where $t\in[0,T]$ is the diffusion timestep, and $\alpha_t \in [0,1]$ is a decaying parameter such that $\alpha_0 = 1$ and $\alpha_T = 0$. 
The diffuser $f_{\theta}$ is trained to predict the $\epsilon$ for various timesteps, by minimizing the objective $\frac{1}{N} \sum_i \mathbb{E}_{t, \epsilon} \mathcal{L}(x_i, t, \epsilon; f_{\theta})$, where $N$ is the training set size, and

\vspace{-0.55cm}
\begin{align}
\mathcal{L}(x, t, \epsilon; f_{\theta}) = \lVert \epsilon - f_{\theta}(x_t, t) \rVert_2^2\text{.}
\label{eq:pixel_training_loss}
\end{align}
\vspace{-0.55cm}

The generation iteratively removes the noise prediction $f_{\theta}(x_t, t)$ from $x_t$ for $t=T,T-1,...,0$, starting from $x_T\sim\mathcal{N}(0,I)$, and obtaining a generated image $x_{t=0}$. To guide this process, for conditional image generation the diffuser $f_\theta$ receives an additional input $y$, which represents a class label for the class-conditional \DMs \cite{ho2020denoising} 
or a text embedding, obtained from a pretrained language encoder like CLIP~\citep{radford2021learning}, for the text-to-image \DMs~\citep{rombach2022high,ramesh2022hierarchical,saharia2022photorealistic}.

Latent diffusion models~\citep{rombach2022high} (LDMs) improve \DMs by conducting the diffusion process in the latent space, which significantly reduces computational complexity, making training scalable and inference more efficient. For the LDMs, the encoder $\mathcal{E}$ transforms the input $x$ to the latent representation $z=\mathcal{E}(x)$ and Equation~\ref{eq:pixel_training_loss} becomes 
\vspace{-0.55cm}

\begin{align}
\mathcal{L}(z, t, \epsilon; f_{\theta}) = \lVert \epsilon - f_{\theta}(z_t, t) \rVert_2^2.
\label{eq:training_loss}
\end{align}

\vspace{-0.55cm}
\paragraph{Membership Inference Attacks.}
MIAs aim to determine whether a specific data point was used to train a given machine learning model~\citep{DBLP:conf/sp/ShokriSSS17}.
Extensive research has explored MIAs against supervised machine learning models~\citep{carlini2022membership,sablayrolles2019white,truex2019demystifying,yeom2018privacy}. 
On the high level, MIAs operate on the premise of overfitting, assuming that training data points (members) exhibit smaller training loss
compared to data points not encountered during training (non-members).
Initial MIAs against \DMs~\citep{carlini2021extracting}
focus on assessing the membership of samples by evaluating the model's noise prediction loss.
Their findings establish that the loss value at the \textit{diffusion timestep} $t=100$ proves most discriminative between member and non-member samples.
Intuitively, if $t$ is too small ($t<50$), the noisy image resembles the original, making the noise prediction too easy. Otherwise, if $t$ is too large ($t>300$), the noisy image resembles random noise, making the task overly challenging.
Among the recent MIA approaches targeting \DMs, the Step-wise Error Comparing Membership Inference (SecMI) attack~\citep{duan23bSecMI} infers membership by estimating errors between the sampling and inverse sampling processes applied to the input $x$ also at timestep $t=100$.
Following the same overfitting principle, the Proximal Initialization Attack (PIA)~\citep{kong2024an} enhances SecMI by assessing membership based on the difference in the model's noise prediction for a clean sample $x$ at timestep $t=0$ and a noised sample $x_t$ at $t=200$, where the method was found to be most discriminative.

\vspace{-0.3cm}

\paragraph{Protecting Intellectual Property in DMs}
 Protecting intellectual property (IP) in \DMs involves safeguarding against unauthorized usage of trained models and attributing generated data to their source models, while also protecting the IP of the data used for training. %
 Several attribution methods focus on watermarking at both the model and input levels, embedding invisible watermarks into generated images or subtly influencing the sampling process to create model fingerprints~\citep{Fernandez_2023_ICCV,liu2023mirror}. Other techniques explore fingerprinting methods, where unique patterns or signals are embedded into generated data for identification purposes~\citep{NEURIPS2023_b54d1757, yu2021artificial}. However, those methods protect the IP in trained models and generated data, leaving the IP of \textit{training data} out of scope. To solve this issue, various approaches aim to protect against style mimicry and unauthorized data usage by adding perturbations to images or detecting unauthorized data usage through injected memorization or protective perturbations~\citep{golatkar2024training,291164,Van_Le_2023_ICCV,wang2024simac,wang2024diagnosis,xue2024toward}. However, existing methods have important drawbacks, such as limiting the data usage for consensual applications and providing no protection if the data IP has already been breached. Moreover, a malicious party may attempt to overcome the safety mechanism by image purification methods~\cite{cao2023impress}. 
 Our proposed method fills in those gaps by enabling data owners to identify whether their data has been illegitimately used for training, without any requirements to modify the protected content.
 While the previous work showed the possibility of computing the influence of the training data points on the generated outputs~\citep{zheng2024intriguingPropertiesDM}, we propose to go a step further and exactly detect which data points are used for training.

\section{Limitations of MIAs in Member Detection}
\label{sec:mia_limit}

We rigorously evaluate existing MIAs to test their ability to detect training members in large, complex DMs. Prior studies~\citep{duan23bSecMI, kong2024an} reported success with MIAs in accurately identifying \DMs training members; however, these results were often based on toy models or datasets (\eg CIFAR100~\cite{Krizhevsky09learningmultiple}) that do not reflect the complexity of high-dimensional, diverse DM setups. Our analysis on state-of-the-art DMs trained on extensive datasets (\ie ImageNet-1k~\cite{deng2009imagenet} or COCO~\cite{veit2016coco}) reveals significant performance limitations of existing MIAs and key factors that contributed to overestimated effectiveness in previous work. Full details are provided in~\cref{app:mia_issues}.

\subsection{Evaluated MIAs}
\label{sec:mia_features}

\begin{enumerate}
    \item \textit{Denoising Loss}~\citep{carlini2023extracting}: The loss is computed from Equation~\ref{eq:training_loss} five times for the diffusion timestep $t=100$, as indicated in the original paper. The final membership score is the average loss, where a lower value indicates that the sample is a member.

    \item \textit{SecMI\textsubscript{stat}}~\citep{duan23bSecMI}: The membership score extracted by SecMI aims to approximate the posterior estimation error of $f_{\theta}$ on the latent $z$ (obtained from the image encoder part), claiming it should be lower for members than for non-members.
    
    \item \textit{PIA}~\citep{kong2024an}: The score extracted by PIA aims at capturing the discrepancy between the noise prediction on a clean sample's latent
    $z$ and the noise prediction on its noised version $z_t$ at time $t$. This discrepancy should be lower for members.
    
    \item \textit{PIAN}~\citep{kong2024an}: This MIA is an adaptation of the original PIA to further strengthen the membership signal. The noise prediction on $z$ is normalized, so it follows the Gaussian distribution. Similar to PIA, the scores returned from PIAN are expected to be lower for members than for non-members.
\end{enumerate}

\subsection{Experimental Setup}
\label{sec:mia-setup}

\paragraph{Models.} We evaluate class-conditioned, as well as text-conditioned state-of-the-art \DMs of various architectures, namely LDM~\citep{rombach2022high}, U-ViT~\citep{bao2023all}, and DiT~\citep{peebles2023scalable}. We employ already trained checkpoints provided by the respective papers~\citep{LatentDiffusionModelsCodebase, UVitCodebase, DiTCodebase}.
For class-conditioned generative tasks, LDM offers one model checkpoint with the resolution of 256x256 %
(LDM256). For U-ViT and DiT, we have access to models operating on resolutions of 256x256 and 512x512 (U-ViT256, U-ViT512, DiT256, DiT512). Additionally, we conduct experiments on text-conditioned models based on U-ViT architecture (U-ViT256-T2I, U-ViT256-T2I-Deep) and a newly trained unconditional U-ViT256-Uncond model (see~App.~\ref{app:experimental_setup}). 

\paragraph{Datasets.} For the class-conditional evaluation, we use models trained on ImageNet-1k~\citep{deng2009imagenet}. This dataset contains large-sized colored images with 1000 classes. There are 1,281,167 training images and 50,000 test images. For text-conditional task evaluation, we use models trained on COCO-Text dataset~\citep{veit2016coco}, a large-scale object detection, segmentation, and captioning dataset which contains 80,000 training images and 40,000 test images, each with 5 captions.

\subsection{Performance of MIAs in Member Detection}
\label{sec:mia_eval_alt}
Our results indicate that the existing MIAs achieve performance comparable to random guessing. 
We present the aggregated max and average \textbf{TPR@FPR=1\%} across 8 DMs in~\cref{tab:mia_tpr_avg}, and defer the full evaluation to~App.~\ref{app:mias_eval}.
For completeness, we provide AUC (Table~\ref{tab:mia_auc_app}), accuracy (Table~\ref{tab:mia_acc_app}), and ROC curves (Fig.~\ref{fig:mia_rocs}) of MIAs there.

\begin{table}[h!]
\scriptsize
\centering
\caption{\textbf{TPR@FPR=1\% for MIAs.} Performance of existing MIAs in identifying training members is limited.}
\begin{tabular}{ccc}
\toprule
\textbf{Attack} & \textbf{Max TPR@FPR=1\%} & \textbf{Mean TPR@FPR=1\%} \\
\midrule
Denoising Loss~\citep{carlini2022membership} & 2.24 & 1.61 \\
SecMI$_{stat}$~\citep{duan23bSecMI} & 2.44 & 1.50 \\
PIA~\citep{kong2024an} & 5.57 & 2.18 \\
PIAN~\citep{kong2024an} & 1.53 & 1.03 \\
\bottomrule
\end{tabular}
\label{tab:mia_tpr_avg}
\vspace{-0.24cm}
\end{table}

\section{Our \ours Method}
\label{sec:method}

Recognizing the limitations of MIAs on large, state-of-the-art DMs, we shift our focus to DI and introduce our \ours{} method. To achieve reliable and confident detection of data collections used in model training, we go beyond simply aggregating features from existing MIAs. 
Our \ours consists of four stages: (1) data and model preparation, (2) feature engineering and extraction, extended by our three newly proposed detection methods (3) a scoring function that maps these features to scores, and (4) a rigorous statistical hypothesis testing, enabling high-confidence decisions. We visualize and describe \ours in \Cref{fig:schema}.

 \vspace{-0.075cm}
\paragraph{Dataset Inference.}

DI was initially introduced as a tool for detecting model stealing attacks~\citep{tramer2016stealing}. 
In the context of supervised models~\citep{maini2021dataset}, DI involves crafting features for a set of training data points, inputting them into a binary classifier, and applying statistical testing to establish model ownership. 
The features of supervised learning are based on the fact that classifiers are trained to maximize the distance of training examples from the model’s decision boundaries while test examples typically lie closer to these boundaries, as they do not influence the model's weights during training.
DI was extended to self-supervised learning (SSL)~\citep{dziedzic2022dataset} by observing that training data representations exhibit a markedly different distribution from test data representations.
 Building on this intuition, 
 we design specific features based on the DM's behavior
 for a set of data points that we want to test for potential (illegitimate) use in training the DM. We then map those features to scores on which we apply statistical testing. Unlike traditional DI, which focuses on ownership resolution for the entire model, our approach is tailored for data verification, allowing owners of small subsets of the \DM's training data to verify their use in model training.

 \vspace{-0.075cm}
\paragraph{Notation and Setup.} 

We denote \P as a set of samples that we suspect to be (illegitimately) used for training the \DM. Those are \textit{published} samples provided by the data owner who wants to make a claim for their intellectual property.
Further, we refer to \U as a set of \textit{unpublished} samples, from the same distribution as \P, that serves as the validation set. In real scenarios, \U might come from a creator's unpublished data or sketches of their released work. We assume \P to be \textit{i.i.d.} with U. 

\vspace{-0.2cm}
\paragraph{Data Preparation and Processing.}
We split \P and \U into \Pctrl, \Ptest, and \Uctrl, \Utest.
We extract the final full set of features for \Pctrl and \Uctrl and train the \scoring $s$ to tell apart members from non-members, such that $s$ eventually outputs higher values when presented with a member. Then, we apply $s$ to the features extracted from \Ptest and \Utest. Finally, we perform statistical testing to find whether the scores returned by $s$ on \Ptest are significantly higher than those on \Utest, which would indicate that \P was, indeed, used to train the \DM.

\vspace{-0.2cm}
\paragraph{Threat Model.} We design \ours as a tool for use in legal proceedings. Consequently, the \ours procedure is carried out by a third trusted party, referred to as an \textit{arbitrator}.  The arbitrator is approached by a victim, whose private data might have been potentially used in a training of a \DM. The arbitrator executes \ours either in the gray-box model access, (can only obtain outputs, \ie noise predictions, for given inputs to a \DM at an arbitrary timestep $t$) or in the white-box model access (where \DM's internals and parameters can be inspected). The access type depends on the requirements of the features used in \ours (we provide more details in App.~\ref{app:access-type}).

\subsection{Features}
\label{sec:our_features}

We utilize MIAs (\cref{sec:mia_features}) as the source of features for \ours. Additionally, to increase the discriminative capabilities of our \ours, we propose the following three novel features that can be extracted from a \DM to provide additional information on a sample's membership score. Our final feature extractor implements a function $f_{e}: \gR^{C\times H\times W}\rightarrow\gR^{k}$,  with $C$, $H$, $W$ denoting the channels, height, and width of an input sample, respectively, and $k$ being the dimensionality of the extracted feature vector.

\vspace{-0.2cm}
\paragraph{Gradient Masking (GM).} 
This feature aims at capturing the difference in the ability to restore destroyed semantic information between members and non-members.
It is inspired by the \textit{Degrade, Restore, Compare} (DRC) idea from~\citet{fu2024model} who identify that for members, a restoration is more successful. 
To compute the feature, we first capture the gradient $\mathbf{g}=|\nabla_{z_t}\mathcal{L}(z_t, t, \epsilon; f_{\theta})|$. 
Intuitively, $\mathbf{g}$ indicates the influence each feature value in the latent $z_t$ has on the loss $\mathcal{L}$. 
We are interested in the features from $z_t$ that exhibit the highest influence on the loss.
Therefore, we create a binary mask \textbf{M} for the top 20\% values in $\mathbf{g}$. %
This mask indicates significant regions of the latent representation $z_t$.
Next, we obtain $\hat{z_t}=\epsilon\cdot\mathbf{M}+z_t\cdot\mathbf{\neg M}$, with the significant values of $z_t$ destroyed by replacing them with random noise $\epsilon\sim\mathcal{N}(0,I)$, and the rest left unchanged.
Finally, we compute $||(\epsilon-z_t)\cdot\mathbf{M}-f_{\theta}(\hat{z_t},t)\cdot\mathbf{M}||_2^2$ as the feature.
This feature expresses the reconstruction loss over the semantically most relevant regions and should be lower for members.
We calculate the feature at multiple diffusion timesteps $t$, to further strengthen the signal.
Note that we differ in our feature computation from~\citet{fu2024model} in two significant aspects. 
(1) While their DRC employs powerful third-party self-supervised vision encoders like DINO~\citep{caron2021dino} to identify semantically significant regions, we utilize only the information from within the \DM to obtain the mask \textbf{M}. (2) Additionally, we utilize the model's loss as our final signal, instead of computing cosine similarity between representations returned from DINO for clean and restored samples, rendering our method more self-contained and independent of the signal from other models.

\vspace{-0.2cm}
\paragraph{Multiple Loss (ML).}%
To increase the membership signal from the model prediction loss, we compute Eq.~\ref{eq:training_loss} at multiple (10) diffusion timesteps $t=0,100,...,900$ to provide more information to train the scoring function $s$. %

\begin{figure*}[t]
    \centering
    \includegraphics[width=0.8\linewidth, trim=0cm 0cm 0cm 0.2cm, clip]{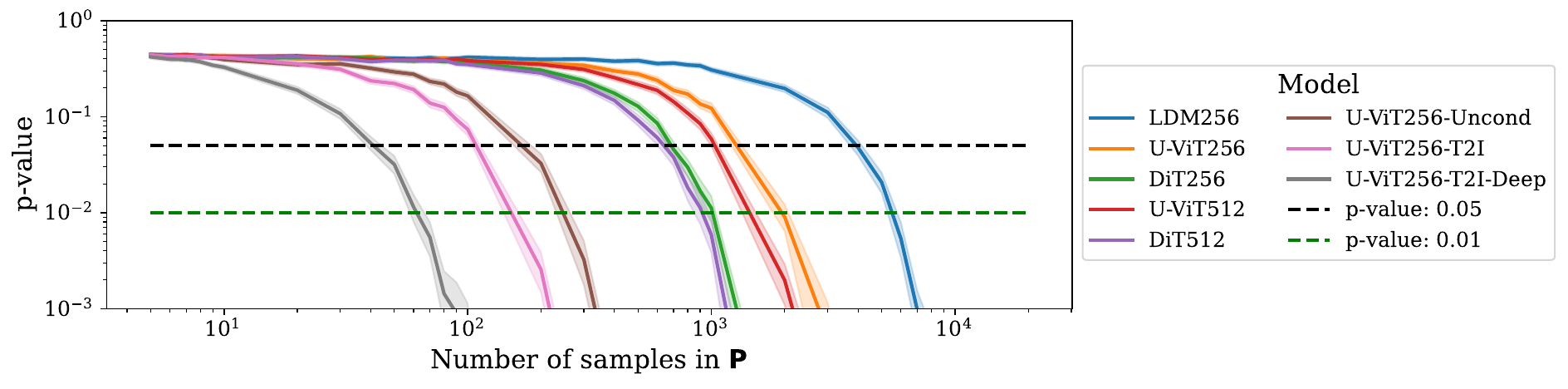}
    \vspace{-0.3cm}
    \caption{\textbf{Results of \ours on various DMs.} Solid lines indicate p-values aggregated over 1000 randomized trials for each size of \P, shaded areas around the lines are 95\% confidence intervals. \ours confidently rejects $H_0$ with as low as 70 suspect samples from the data owner. \ours's performance increases with larger model sizes and smaller training sets.
    }
    \vspace{-0.35cm}
    \label{fig:cdi_results}
\end{figure*}

\vspace{-0.2cm}
\paragraph{Noise Optimization (NO).} %
We leverage an insight initially observed in supervised classifiers, namely that the difficulty of changing the predicted label of a sample through adversarial perturbations differs between members and non-members~\citep{zheng_adv_mia}. 
In particular, it takes a stronger perturbation to change the prediction for members.
The reason is that ML models return more confident predictions on training samples (members).
To craft our feature, we adapt this intuition to \DMs. We note that perturbing a noised sample $z_t$ to minimize the model noise prediction loss expressed in Equation~\ref{eq:training_loss} achieves better results, \ie lower loss values for member samples.
Specifically, we conduct an unbounded optimization of the perturbation $\delta$ applied to the noised latent representation $z_t$ at timestep $t=100$. Our objective function is defined as: 
$\argmin_{\delta} ||\epsilon - f_{\theta}(z_t + \delta, t)||_2^2$. 
To optimize this objective, we employ the 5-step L-BFGS algorithm~\citep{LBGFS} (which is commonly used to generate adversarial perturbations~\citep{carlini2017towards, szegedy2013intriguing}).
We use the resulting values of the noise prediction error $||\epsilon - f_{\theta}(z_t + \delta, t)||_2^2$ and the amount of perturbation $||\delta||_2^2$ as features.

We provide further analysis of our features in~App.~\ref{app:our_features_analysis}.

\subsection{Scoring Model}
\label{sec:method_scoring}

Based on the set of features extracted for \Pctrl and \Uctrl, we train a logistic regression model, $s: \mathbb{R}^k\rightarrow[0,1]$. 
We then apply $s$ to the features extracted for \Ptest and \Utest and use the resulting logits $s(f_e(\Ptest))$ and $s(f_e(\Utest))$ as membership confidence scores, with higher values referring to higher confidence that the given input sample is a member. 

The motivation behind relying on the feature vector extracted by $f_e$ is that a single feature-based score is prone to high variance~\citep{carlini2022membership}, in turn making it more difficult to perform successful data detection. 
In contrast, combining multiple features should amplify the signal and improve the performance. 
Our experiments confirm this intuition, as we show in Sec.~\ref{sec:mia_eval_alt} and~\ref{sec:ours_eval}.
Moreover, the scoring model addresses the challenge of determining which features provide the strongest membership signal for a given model. By aggregating information across multiple features, $s$ highlights the most relevant signals for detecting membership.

\subsection{Statistical Testing}
\label{sec:method_stats}
Finally, we perform a two-sample single-tailed Welch's t-test. Our null hypothesis expresses that mean scores for \Ptest are not significantly higher than the ones for \Utest, \ie $H_0: \overline{s(f_e(\Ptest))}\leq \overline{s(f_e(\Utest))}$, where $\overline{s(f_e(\Ptest))}$ and $\overline{s(f_e(\Utest))}$ are the mean of scores returned from $s$ on the features of \Ptest and \Utest, respectively.
Rejecting $H_0$ at a significance level $\alpha=0.01$, \ie obtaining p-value $<0.01$, confirms that samples from \Ptest has been (illegitimately) used to train the queried \DM. 

Our choice of such a low $\alpha$ parameter is motivated by the \textbf{TPR@FPR=1\%} metric established for MIAs~\citep{carlini2022membership}, based on the intuition that the false positives are more harmful than false negatives in real-world applications, \eg court cases.
To improve the soundness of our statistical tests, we perform \ours 1000 times on randomly sampled subsets of \P and \U, and aggregate the obtained p-values~\citep{kost2002combining,vovk2019combining} (we provide more details in App.~\ref{app:pvalues}).

\section{Empirical Evaluation}
\label{sec:empirical_evaluation}

\begin{table*}[h!]
\scriptsize
\centering
\caption{\textbf{Impact of the statistical testing}. 
The values in the table are \textbf{TPR@FPR=1\%} and are in \%. Results represent the \textit{set-level} MIA (without the statistical testing) vs \ours with the statistical testing.  The size of \P is 1000. Statistical testing is essential for \ours.}
\begin{tabular}{ccccccccc}
\toprule
 & \textbf{LDM256} & \textbf{U-ViT256} & \textbf{DiT256} & \textbf{U-ViT512} & \textbf{DiT512} & \textbf{U-ViT256-Uncond} & \textbf{U-ViT256-T2I} & \textbf{U-ViT256-T2I-Deep} \\
\midrule
Set-level MIA (no t-test) & 10.20 & 22.90 & 10.50 & 6.50 & 0.00 & 33.40 & 23.20 & 32.50 \\
\textbf{\ours (Ours)}
& \textbf{24.92} & \textbf{62.74} & \textbf{93.00} & \textbf{74.43} & \textbf{93.76} & \textbf{100.00} & \textbf{100.00} & \textbf{100.00} \\
\bottomrule
\end{tabular}
\label{tab:set_mia_vs_cdi}
\end{table*}

\begin{figure*}
    \centering
    \includegraphics[width=0.84\linewidth, trim=0cm 0cm 0cm 0cm, clip]{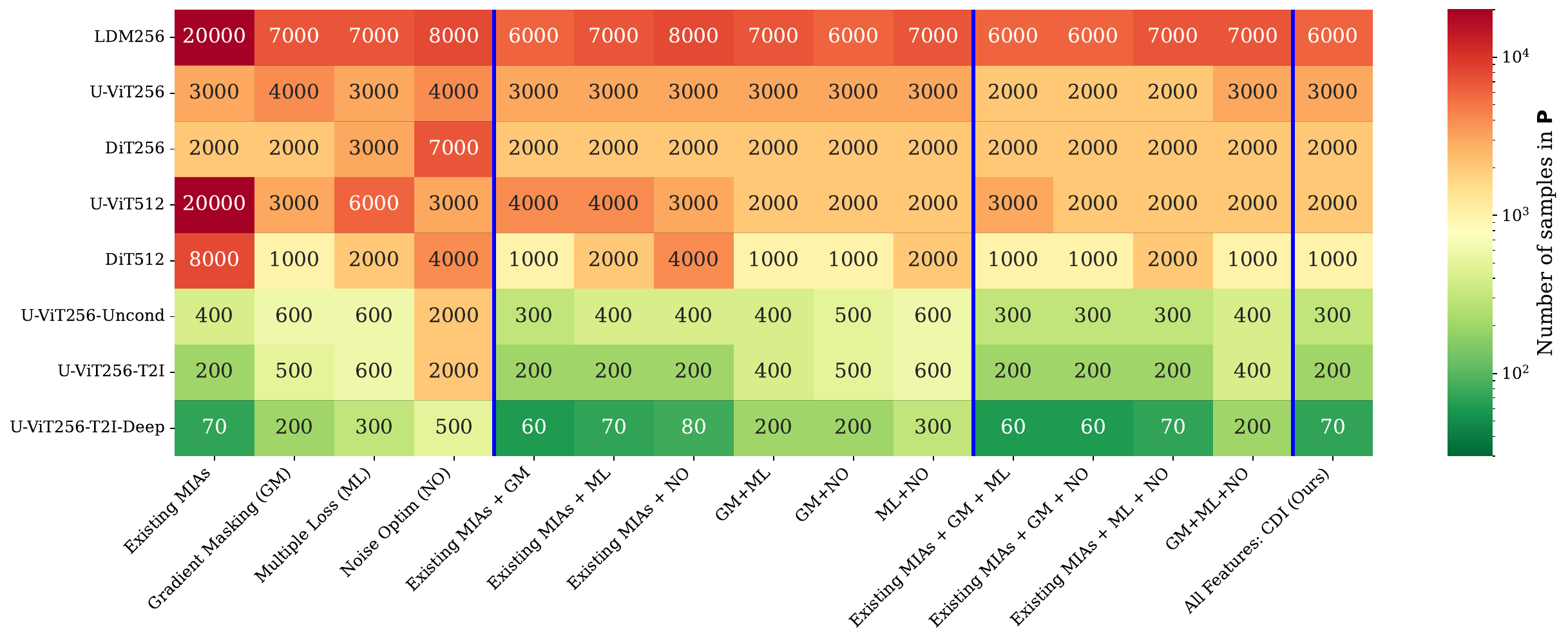}
    \vspace{-0.3cm}
    \caption{\textbf{Impact of feature selection.} The values in the cells indicate the minimum size of \P needed to reject $H_0$. Blue vertical lines separate results by the complexity of the feature set used to fit $s$: 
    (left) our novel individual features from~\cref{sec:our_features} and a joint existing MIAs feature, (second from left) all possible combinations of two, and of three features (second from right), and (right) all available features.
    }
    \label{fig:ablation_main}
    \vspace{-0.5cm}
\end{figure*}

\paragraph{Our CDI Setup.} 
We use the diffusion models and datasets as specified in \cref{sec:mia-setup}.
To instantiate our \ours, we draw samples from the train sets to represent \P and samples from the test sets to represent \U.
We set $|\P|=|\U|$ for all experiments. 
The maximum total size of $|\P|+|\U|$ we use for our experiments is 40,000 samples. Note, that this number is chosen only as a starting point %
and the number of data points for \P and \U  that \ours requires to confidently reject $H_0$ is much lower and depends on the targeted model (see Fig.~\ref{fig:cdi_results}).

To maximize the use of both \P and \U while minimizing the number of samples required for our method, we implement a $k$-fold cross-validation with $k=5$. The features extracted from the public samples (\P) and unpublished samples (\U) are divided into 5 folds. In each iteration, one fold is designated as the test set, containing \Ptest and \Utest features, while the remaining $k-1$ folds form the control set, comprising \Pctrl and \Uctrl features, which are used to train the scoring model $s$. This process is repeated across all splits, ensuring that each sample in \P and \U is used exactly once in the test set as part of \Ptest and \Utest.

This procedure ensures that the statistical testing is performed with $|\Ptest| = |\P|$ and $|\Utest| = |\U|$, allowing us to extract signals that identify training data from the entirety of the \P and \U sets.

\subsection{Our CDI Confidently Identifies Collection of Data Samples as Training Data}
\label{sec:ours_eval}
We summarize the success of our \ours for diverse \DMs and datasets in Fig.~\ref{fig:cdi_results} (following the standard evaluation of DI as proposed in~\citep{maini2021dataset}). %
We report p-values as the confidence in the correct verification for different sizes of suspect data sets \P.
Our results highlight that \ours already enables a confident ($p<0.01$) dataset identification  
with as little as 70 samples (for instance, for U-ViT256-T2I-Deep \DM trained on the COCO dataset) provided by the data owner. 
For \DMs trained on larger datasets like ImageNet, we observe the need to increase the size of \P to confidently reject the null hypothesis. More details on the impact of the size of \P on the confidence of \ours can be found in~App.~\ref{app:ptest_size}.
In general, in our results, we identify the following trends:
(1) For a given \DM architecture, trained on given dataset, the number of samples required for the confident identification of the training data decreases with increasing input resolution (see~App.~\ref{app:insights}). 
(2) The larger the overall training set of the model, the more samples are needed for a confident claim (see also~App.~\ref{app:insights}).
(3) The higher the number of model training steps the stronger the signal for identifying training data as shown in \cref{fig:cdi_results_training_steps} in~App.~\ref{app:ts}.

\subsection{Analysis of the Success of \ours}
We perform multiple ablations on \ours's building blocks to deepen understanding of its success. First, we show the importance of statistical testing as a core component of \ours. Then, we show that our features indeed boost the performance of \ours. Next, we demonstrate that our method remains effective even when not all samples from the suspect set were used in the \DM's training set. 
Finally, we show that \ours does not return false positives. We include additional evaluation of \ours in~App.~\ref{app:cdi_tpr_eval} and \ref{app:mias_di}, 
analysis of scoring model $s$ in App.~\ref{app:xai_lr}, 
and time complexity in App.~\ref{app:compute}. In \Cref{app:extending} we evaluate \ours on additional \DMs and showcase how \ours can be extended with additional feature extraction methods.

\vspace{-0.1cm}
\paragraph{\textbf{Statistical Testing is Crucial for DI.}}
\label{sec:stat_cdi_ablation}

In this ablation study, we assess the impact of removing the t-test from \ours and demonstrate that simply aggregating the MIA results for multiple samples is insufficient to reliably identify data collections used in DM training. To conduct this comparison, we aggregate membership scores across a set of samples to determine if any members are present in the set.
We define a \textit{set membership score} as the highest membership score within a subset, hypothesizing that sets composed of members will yield higher scores than non-member sets. For evaluation, we sample 1000 subsets each from  \U  (non-members) and  \P  (members). We refer to this approach as \textit{set-level MIA}, and execute this procedure for scores obtained from the scoring model component of \ours. We report \textbf{TPR@FPR=1\%} in~\cref{tab:set_mia_vs_cdi}.

A direct comparison between \ours's p-values and \textbf{TPR@FPR=1\%} for set-level MIA is challenging due to the differing metrics. To align them, we compute the power of the t-test with $\alpha=0.01$, which is equivalent to \textbf{TPR@FPR=1\%} (see App.~\ref{app:statistical_testing}). This approach allows us to directly compare set-level MIA to \ours without altering its methodology.  
The results in~\cref{tab:set_mia_vs_cdi} underscore the critical role of statistical testing in \ours. Set-level MIA without statistical testing underperforms, while \ours with its rigorous testing achieves near-perfect performance for most of the models.

\begin{table*}[h!]
    \scriptsize
    \centering
    \caption{\textbf{Robustness of \ours against false positives.}
    We depict averaged p-values returned by our method based on the data used within \P with $|\P|=10000$. We sample $1000$ \P and \U sets. The results show that when testing non-members (both \P and \U contain only nonmembers), we obtain high p-values, significantly above the significance level ($\alpha=0.01$), \ie we cannot reject the null hypothesis and do not identify the data from \P as members. In contrast, when testing with member data points (\P contains members and \U contains nonmembers), our results are always significant, \ie $p<0.01$, and we correctly identify the given set as members.
    }
    \begin{tabular}{ccccccccc}
    \toprule
    \textbf{Data in \Ptest} & \textbf{LDM256} & \textbf{U-ViT256} & \textbf{DiT256} & \textbf{U-ViT512} & \textbf{DiT512} & \textbf{U-ViT256-Uncond} & \textbf{U-ViT256-T2I} & \textbf{U-ViT256-T2I-Deep} \\
    \midrule
    \textbf{Members} & $10^{-6}$ & $10^{-21}$ & $10^{-59}$ & $10^{-31}$ & $10^{-66}$ & $10^{-266}$ & 0.00 & 0.00 \\
    \textbf{Non-members} & 0.40 & 0.39 & 0.39 & 0.39 & 0.40 & 0.40 & 0.39 & 0.38 \\
    \bottomrule
    \vspace{-0.2cm}
    \end{tabular}
    \label{tab:robust_fp}
\end{table*}

\vspace{-0.1cm}
\paragraph{\textbf{Our Novel Features Significantly Decrease the Number of Samples Required for Verification.}}
Our results in~\cref{fig:ablation_main} indicate that introducing our new features improves the efficiency of \ours substantially in comparison to the joint features extracted by the MIAs from~\cref{sec:mia_features}, in the following referred to as \textit{existing MIAs}. 
Applying our new features in \ours leads to a remarkable reduction of the number of samples needed to reject the null hypothesis with $p<0.01$, especially in the initially most challenging cases of models trained on higher-resolution large datasets.
For instance, for U-ViT512 the number of samples required from a data owner for confident verification decreases from 20000 to 2000.
This makes \ours more practical and applicable to more users who have smaller datasets that they would like to verify.

Regarding our novel features, the most influential one is GM. We note that utilizing only existing~MIAs~+~GM, we are able to obtain performance very close to \ours. 
The extension of the feature set only with NO yields the smallest improvement over the alternatives in most cases, however, discarding it entirely (see: existing~MIAs~+~GM~+~ML vs All Features) results in worse performance in the case of U-ViT512, highlighting the need for a diverse source of the signal to obtain confident predictions. Our ML feature succeeds in capturing a better membership signal than its existing~MIAs-based counterpart (Denoising Loss), striking a middle ground between GM and NO.

\begin{figure}[t!]
    \centering
    \includegraphics[width=1.0\linewidth, trim=0.3cm 0.3cm 0.2cm 0cm, clip]{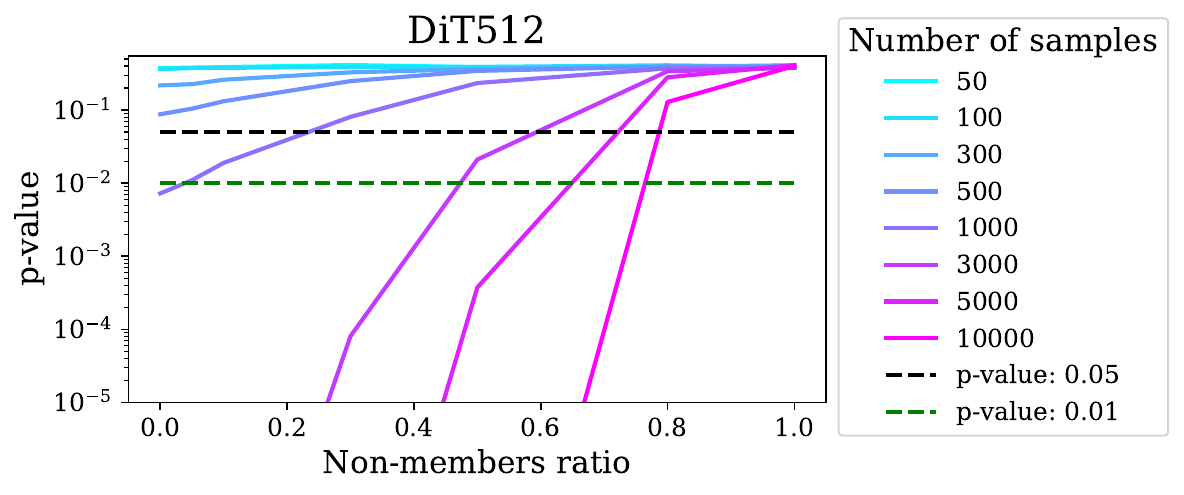}
    \caption{\textbf{Impact of non-members ratio in \P on \ours}. The lines represent p-values for a given non-member ratio while varying sizes of \P. 
    }
    \label{fig:noise_influence}
\vspace{-0.2cm}
\end{figure}

\vspace{-0.075cm}
\paragraph{\textbf{CDI is Effective Even When Not All Data Samples Were Used as Training Data.}}
We investigate \ours’s behavior in cases where only a part of the samples in the suspect set \P was used to train the \DM, \ie \P contains a certain ratio of non-members, while the remaining samples in \P are members. Practically, this corresponds to the situation where a data owner has a set of publicly exposed data points and suspects that all of them might have been used to train the \DM, whereas, in reality, some were not used. This can happen, \eg due to internal data cleaning on the side of the party who trained the \DM. In particular, the data owner does not know which of their samples and how many of them have not been included into training.
In Fig.~\ref{fig:noise_influence} (extended by Fig.~\ref{fig:noise_influence_full} in~App.~\ref{app:contamination}), we present the success of \ours under different ratios of non-member samples in \P. Note that our evaluation is the result of 1000 randomized experiments for each non-members ratio, model, and \P size. We observe that \ours remains effective when the non-member ratio is over 0.5 and 0.8 for some models, \ie it still correctly identifies \P as training data of the model. Overall \ours's robustness is higher when the data owner provides larger suspect datasets \P. This is reflected in the p-value at the same non-member ratio decreasing as the number of samples in \P increases.

\vspace{-0.075cm}
\paragraph{\textbf{CDI is Effective Even Under Gray-box Model Access.}}We analyze the effectiveness of performing \ours in the gray-box model access scenario, as defined in the threat model (\cref{sec:method}). Therefore, we include only the original MIA features and Multiple Loss in \ours. 
Even in this case, \ours remains effective under gray-box model access and can reject the null hypothesis.
In this more difficult scenario, across the eight tested \DMs, \ours requires on average \textbf{one-third} more samples in \P compared to the white-box access (where all features can be used). We refer to App.~\ref{app:gray-vs-white} for a detailed comparison. 

\vspace{-0.075cm}
\paragraph{\textbf{CDI is Robust Against False Positives.}}
While \ours can correctly identify suspect datasets even when not all samples were used as training data, it raises the concern of false positives, \ie reporting data as used for training a \DM when it was not.
In particular, \ours should only reject the null hypothesis if \P contains (some) members and yield inconclusive results otherwise.
To show that \ours is robust against false positives, 
we instantiate \P only with non-member samples. %
Our results in Table~\ref{tab:robust_fp} highlight \ours’s reliability in distinguishing between non-member and member sets without false positives.

\section{Conclusions}

We introduce \ours as a method for data owners to verify if their data has been (illegitimately) used to train a given \DM. While existing MIAs alone are insufficient to confidently determine whether a specific data point was used during training, \ours overcomes this limitation. By selectively combining features extracted from MIAs with novel handcrafted features and applying them across a larger data set, we achieve a reliable discriminator for identifying datasets used in \DM training. Our rigorous feature engineering amplifies the signal in \ours, 
enabling individual artists even with smaller collections of art to benefit from our method.

\subsubsection*{Acknowledgments}
This work was supported by the German Research Foundation (DFG) within the framework of the Weave Programme under the project titled "Protecting Creativity: On the Way to Safe Generative Models" with number 545047250.  This research was also supported by the Polish National Science Centre (NCN) grant no. 2023/51/I/ST6/02854 and 2020/39/O/ST6/01478 and by the Warsaw University of Technology within the Excellence Initiative Research University (IDUB) programme.
Responsibility for the content of this publication lies with the authors.

{
    \small
    \bibliographystyle{ieeenat_fullname}
    \bibliography{main.bib}
}

\appendix
\clearpage
\section{Broader Impact}
\label{app:broader_impact}

Our research addresses the pressing issue of verifying the ownership of data used for training large image generation models, as highlighted by recent legal disputes \cite{gety_v_stability}. By introducing \ours, we aim to enable data owners to verify if their data was (illegitimately) used for training \DMs. Our work contributes to a more transparent and accountable ML ecosystem, aligning with broader societal values of fairness and respect for data ownership. We anticipate \ours will have a positive impact on both the ML community and society, promoting responsible and fair development of ML models.
We make our code available at {\url{https://github.com/sprintml/copyrighted_data_identification}}.

\section{Limitations}
\label{app:limitations}

In this paper, we assess the effectiveness of \ours for image generation \DMs. Although we believe that our methodology extends well to other modalities such as text or video, we do not perform an experimental evaluation in these areas.
Our research focuses on diffusion models as the current state-of-the-art image generators widely employed in commercial applications. While acknowledging the existence of alternative generative image models, which have recently been shown to demonstrate comparable capabilities \citep{kang2023gigagan, pmlr-v202-sauer23a}, we choose to focus on \DMs within the scope of our research.
We note that CDI requires at least gray-box access, i.e. \DM's predictions at an arbitrary timestep $t$ to be effective. This limitation stems from the lack of reliable, strictly black-box MIA methods for \DMs, \ie methods that leverage only the final generated image, while avoiding the pitfalls described in~App.~\ref{app:mia_issues}.

\section{Additional Background}
\label{app:background}

\subsection{Previous Membership Inference Attacks against DMs}
We present the previous MIAs against \DM that provide the initial set of features we use for our \ours.

\label{app:mias}%
\paragraph{Denoising Loss by~\citet{carlini2023extracting}.} The common intuition for utilizing the loss function of the model is that its values should be lower for the training set (members) than validation or test set (non-members). Formally, we follow LiRA~\citep{carlini2022membership}, and its extension to DMs~\citep{carlini2021extracting}, and for each sample, we compute $\lVert \epsilon - f_{\theta}(z_t, t) \rVert_2^2$ at $t=100$. Note, that $z_t$ is obtained by adding $\eps\sim\mathcal{N}(0,I)$ to the original $z$, a process which, by its stochasticity, introduces noise to obtained scores.~\citet{carlini2021extracting} suggest to address this issue by computing the loss for five $z_t$ noised using different $\eps$. The final feature is the mean of these five measurements.

\paragraph{SecMI\textsubscript{stat} by~\citet{duan23bSecMI}.} The feature is based on the assumption that the effect of the denoising process should restore member samples better than non-member samples.~\citet{duan23bSecMI} formalizes this idea by introducing \textit{t-error}, a metric that aims to approximate the estimation error of $f_{\theta}$ on a given $z$. More specifically, \textit{t-error} is defined as 
$\lVert\psi_{\theta}(\phi_{\theta}(\tilde{z}_t,t),t)-\mathbf{\Phi_{\theta}}(z_0,t)\rVert_2^2$, 
where $\psi_{\theta}(z_t,t)=z_{t-1}=\sqrt{\bar{\alpha}_{t-1}}\hat{f}_{\theta}(z_t,t)+\sqrt{1-\bar{\alpha}_{t-1}}f_{\theta}(z_t)$ is the DDIM~\citep{song2020denoising} denoising step,
$\phi_{\theta}(z_t,t)=z_{t+1}=\sqrt{\bar{\alpha}_{t+1}}\hat{f}_{\theta}(z_t,t)+\sqrt{1-\bar{\alpha}_{t+1}}f_{\theta}(z_t,t)$, is the DDIM sampling inverse step, $\hat{f}_{\theta}(z_t,t)=\frac{z_t-\sqrt{1-\bar{\alpha}_t}f_{\theta}(z_t,t)}{\sqrt{\bar{\alpha}_t}}$, $\bar{\alpha}_t=\prod_{k=0}^t\alpha_k$, and $\mathbf{\Phi_{\theta}}(z_s,t)=\phi_{\theta}(\cdots\phi_{\theta}(\phi_{\theta}(z_s,s),s+1),\cdots,t-1)$ is the deterministic reverse. \textit{t-error}, intuitively, should hold lower values for member samples.

\paragraph{PIA~by~\citet{kong2024an}.} PIA builds on the notion that, under DDIM sampling settings, given $z$ and any $z_t$, one can determine the \textit{ground truth trajectory} consisting of intermediate $z_s$, $s\in(0,t)$. Subsequently, $f_{\theta}$ learns to reflect this trajectory and is more competent in it for members. Capturing that difference can act as a membership signal, defined as $\lVert f_{\theta}(z,0)-f_{\theta}(\sqrt{\bar{\alpha_t}}z+\sqrt{1-\bar{\alpha_t}}f_{\theta}(z,0),t)\rVert_5$, where $\bar{\alpha_t}=\prod_{k=0}^t\alpha_k$, and $t=200$. The feature should be lower for members.

\paragraph{PIAN by~\citet{kong2024an}.} An important consideration in PIA is that $f_{\theta}(z,0)$ follows a Gaussian, \ie $f_{\theta}(z,0)\sim\mathcal{N}(0,I)$, as it should make the attack more performant. To assure that,~\citet{kong2024an} propose PIAN (normalized PIA), as an extension of their method. Formally, $\hat{f_{\theta}}(z,0)=C\cdot H\cdot W\sqrt{\frac{\pi}{2}}\frac{f_{\theta}(z,0)}{\lVert f_{\theta}(z,0)\rVert_1}$ with $C, H, W$ denoting the channels, height, and width of the input sample, respectively. However, our experiments in~App.~\ref{app:mias_eval} and~\cref{sec:ours_eval} indicate that this intuition does not translate to latent DMs, which is in line with findings from the original paper~\citep{kong2024an}. %

\begin{figure*}[t!]
\centering
\includegraphics[width=0.9\linewidth]{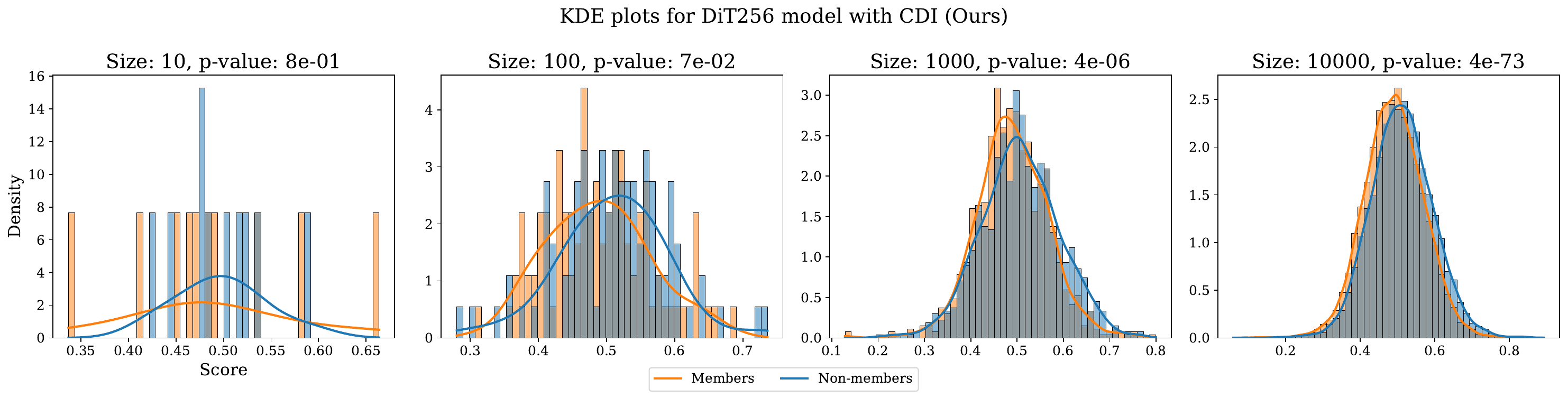}
\vspace{-0.2cm}
\caption{\textbf{KDE plots for varying $|$\Ptest$|$}. We use the DIT256 model.}
\label{fig:size_matters}
\vspace{-0.2cm}
\end{figure*}

\section{On the Mismatch in MIAs Results}
\label{app:mia_issues}

Contrary to the promising results of SecMI\textsubscript{stat}, PIA, and PIAN, our evaluation in~App.~\ref{app:mias_eval} shows that these attacks fail to reach performance significantly higher than random guessing. The following is a methodological analysis of issues in the experimental setup proposed in~\citep{kong2024an, duan23bSecMI, fu2024model}. We identify three pitfalls in their experimental settings: (1) usage of toy models, (2) overfitting to the evaluation set, and (3) distribution mismatch between members and non-members sets.

\subsection{Pitfall 1: Toy Models}

MIAs performance is directly correlated with the level of overfitting in the attacked model~\citep{DBLP:conf/sp/ShokriSSS17, salem2019ml}. Unfortunately, it is a common approach to evaluate the MIAs against \DMs on very small toy models, trained or fine-tuned on small-scale datasets like CIFAR10~\citep{krizhevsky2009learning}, as it speeds up the inference. This in turn can lead to elevated performance which is further reported in published works~\citep{duan23bSecMI, kong2024an, fu2024model}, \eg the \textbf{TPR@FPR=1\%} at the level of 10\% for SecMI\textsubscript{stat} and 30\% for PIAN. We argue that evaluating any type of MIA in such setting is flawed, and provides incorrect insight on the performance of the proposed MIA.

In contrast to previous works, we focus only on evaluating success of \ours on non-overfitted large \DMs from official repositories, trained on high-scale datasets like ImageNet. 

\subsection{Pitfall 2: Overfitting to the Evaluation Set}

SecMI\textsubscript{NNs}~\citep{duan23bSecMI} in its official implementation uses the evaluation set for selection of the best-performing classifier\footnote{\url{https://github.com/jinhaoduan/SecMI/blob/main/mia_evals/secmia.py\#L358}}.

\subsection{Pitfall 3: Mismatch in Distribution of Member and Non-Member Sets}

\citet{dubinski2024towards} highlight the importance of members and non-members sets being indistinguishable from each other in order for the results of MIAs being reliable. Indeed, the mismatch between these can be enough for any classification-based method to succeed, without the context of attacked model. More importantly, a simple Loss attack~\citep{yeom2018privacy} also benefits from this pitfall, as out-of-distribution (incorrect non-members) samples usually achieve significantly higher values of model loss objective than in-distribution samples (members, or correct non-members, \eg test set). Unfortunately,~\citet{duan23bSecMI} for evaluation of their SecMI, as well as~\citet{kong2024an}  for PIA on SOTA Stable Diffusion~\citep{rombach2022high} utilize an external dataset, namely COCO~\citep{veit2016coco}, as their non-members set. In effect, it casts doubt on the reported success of their methods.

In our work we avoid this problem by using validation sets of the \DMs as the source of non-members samples.

\section{On Reporting p-values}
\label{app:pvalues}

To ensure the reliability of the statistical test results reported for \ours, we adopt the following approach. We repeat the t-tests 1000 times on features obtained from randomly sampled subsets of \P and \U, aggregating the resulting p-values. We consider various p-value aggregation methods as reviewed by \citep{kost2002combining,vovk2019combining}, and make our final decision based on the specific context in which we apply \ours.

In our framework, we assume that \ours is executed by an arbitrator approached by a victim whose private data might have been used in the training of a \DM. Each execution of the t-test represents data verification for a \textit{single} data owner, with each p-value corresponding to the test outcome for an \textit{individual case}. To appropriately represent the performance of our method in this setting, we report p-values aggregated using an arithmetic mean over the success for all data owners.

This aggregation method is a conservative approach for reporting our results (comparing to \eg harmonic mean). This is due to the vulnerability of the arithmetic mean to outliers as the p-values are strictly positive, and the mean can be saturated easily by a single large value. In contrast, harmonic mean can be too easily brought down to almost zero by a single good result which would overstate the success of our method.

\section{Impact of the size of the \Ptest on \ours confidence}
\label{app:ptest_size}
We present a visualization of the impact of the size of the \Ptest on the confidence of \ours in~\cref{fig:size_matters}. We sample $n=10,100,1000,10000$ samples randomly from \Ptest and \Utest, and apply \ours. We observe that when we increase the size of \Ptest the method becomes more stable and more confident. This is an inherent feature of statistical testing, a core element of \ours.

\section{Additional Features}

As noted previously, Latent Diffusion Models represent the state-of-the-art in high-resolution image generation for \DMs. Such models are two-stage architectures, where the diffusion process occurs in the latent space of an autoencoder. This observation introduces another potential angle for MIA on latent \DMs. We note that \ours can be extended by incorporating signal from features specifically tailored for the autoencoder part of the model. However, it is important to recognize that the diffusion backbone and the autoencoder are separate models, which can be trained on different datasets. This necessitates caution when deciding whether to incorporate features extracted from the encoder. Nonetheless, in cases where both the autoencoder and the \DM are trained on the same dataset, we claim that the performance of \ours can be further boosted by incorporating membership signals from the autoencoder. To this end, we propose an additional feature that can be utilized in such scenarios.

\textbf{Autoencoder Reconstruction Loss (ARL).} The goal of this feature is to extract differences in the autoencoder reconstruction errors between members and non-members, where members should exhibit significantly lower errors.
To obtain the features, we first note that the current state-of-the-art \DMs consist of the pixel and latent spaces~\citep{rombach2022high}.
For our ARL feature, we leverage the two-stage structure of \DMs and extract the membership signal directly in the pixel space, which contains an autoencoder with the encoder part $\gE$ and decoder $\gD$.
Given an input image $x$, the encoder $\gE$ encodes $x$ into a latent representation $z=\gE(x)$. The decoder $\gD$ reconstruct the image from the latent $z$, yielding $\tilde{x} = \gD(z) = \gD(\gE(x))$.
The autoencoder reconstruction loss serves as the ARL feature computed as $||x-\gD(\gE(x))||_2^2$.

However, for all the models on which we perform our experiments, the autoencoder was trained on different dataset than the diffusion backbone. The LDM model utilizes VQ-VAE\citep{vqvae} trained by \citet{rombach2022high} on the Open Images Dataset V4 \citep{openimages} dataset. All other models use KL autoencoder \citep{vae1, rombach2022high} provided by Stablility AI \citep{vae_checkpoint}. This model was first trained on the Open Images Dataset V4 and then finetuned on subsets of LAION-Aesthetics \citep{schuhmann2022laion5b} and LAION-Humans \citep{schuhmann2022laion5b} datasets. To keep compatibility with existing models trained by Stability AI, only the decoder part was finetuned. Accounting for the difference between the underlying autoencoder and diffuser training datasets present in all models on which we evaluate \ours we do not employ the ARL feature in our framework in this paper. However, we note that it constitutes another source of membership signal applicable in cases when both stages of the latent \DM were trained on the same dataset.  

\section{Extending \ours}
\label{app:extending}
\ours provides an inherently flexible framework for identifying data collections used in the training of \DMs. In this section, we discuss how \ours can be extended with additional feature extraction methods and applied to a broader range of \DM architectures and data modalities.

\subsection{Extending \ours with features from novel MIAs} As the field advances, \ours can be extended with additional feature extraction methods based on novel MIAs. We demonstrate this by incorporating CLiD~\citep{clid} as a feature extraction method within \ours.  First, in \Cref{tab:clid_tpr}, we show that the performance of CLiD MIA remains limited for models trained on ImageNet (first five columns). While it significantly outperforms previous MIAs, as compared to \cref{tab:mia_tpr_app}, it still does not allow reliable membership identification for ImageNet-trained models. Note that we do not evaluate CLiD on U-ViT256-Uncond, as it requires both conditional and unconditional inputs to the DM.

\setlength{\tabcolsep}{2.95pt} %
\begin{table}[h!]
    \centering
    \tiny
    \caption{\textbf{ CLiD~\citep{clid} MIA results at a TPR@FPR=1\%}. Values in the table are in \%.}
    \label{tab:clid_tpr}
    \begin{tabular}{cccccccc}
    \toprule
\textbf{Model} & \textbf{LDM} & \textbf{U-ViT256} & \textbf{DiT-256} & \textbf{U-ViT512} & \textbf{DiT512} & \textbf{U-ViT256-T2I} & \textbf{U-ViT256-T2I-Deep} \\ %
\midrule
TPR@FPR=1\% & 1.52 & 2.18 & 2.74 & 2.03 & 2.00 & 10.91 & 24.11 \\ %
    \bottomrule
    \end{tabular}
\end{table}

Next, we employ features from CLiD in \ours{} and present in \Cref{tab:clid_cdi} that it further improves our method. This confirms our premise from \Cref{sec:method_scoring}, that more meaningful features improve our method. It also demonstrates how \ours{} benefits from advancements in MIAs. When we extend \ours with CLiD, the method needs as few as 30 samples to identify data collections used in \DM training.

\setlength{\tabcolsep}{3pt} %
\begin{table}[h]
    \centering
    \tiny
    \caption{\textbf{Effect of extending \ours{} with features extracted using CLiD.} We report the minimal sample size of \P needed to confidently reject the null hypothesis with \ours.}
    \label{tab:clid_cdi}
    \begin{tabular}{cccccccc}
    \toprule
    \textbf{Model} & \textbf{LDM} & \textbf{U-ViT256} & \textbf{DiT256} & \textbf{U-ViT512} & \textbf{DiT512} & \textbf{U-ViT256-T2I} & \textbf{U-ViT256-T2I-Deep} \\
    \midrule
    Base \ours{} & 6000 & 3000 & 2000 & 2000 & 1000 & 200 & 70 \\
    With CLiD & 4000 & 700 & 400 & 2000  & 700 & 50 & 30 \\
    \bottomrule
    \end{tabular}
\end{table}

\subsection{Extending \ours{} to novel DMs}
With new \DM architectures introduced, \ours{} can be applied to a broader range of \DMs. In \Cref{tab:cdi_new_dms}, we evaluate the performance of \ours{} on SiT~\cite{ma2024sit}, MDT~\cite{gao2023masked}, and DiMR~\cite{Liu2024alleviating}, and show that it successfully identifies the data collections used in \DM training.

\setlength{\tabcolsep}{3pt} %
\begin{table}[h!]
    \centering
    \scriptsize
    \caption{\textbf{Performance of \ours{} on additional \DMs.} We report the minimal sample size of \P needed to confidently reject the null hypothesis with \ours.}
    \begin{tabular}{@{}cccccc@{}}
    \toprule
    \textbf{Model} & \textbf{SiT-XL/2} & \textbf{MDTv2-XL/2} & \textbf{MDTv1-XL/2} & \textbf{DiMR-XL/2R} & \textbf{DiMR-G/2R} \\
    \midrule
    $|\P|$ & 300 & 300 & 200 & 2000 & 2000 \\
    \bottomrule
    \end{tabular}
    \label{tab:cdi_new_dms}
\end{table}

\subsection{Extending \ours{} to other data modalities} 

\ours{} framework is inherently flexible and can be applied to data modalities beyond images, such as text or audio. This extension is straightforward due to the use of feature extraction methods that operate in the latent space of \DMs.

\begin{table*}[h!]
    \scriptsize
    \centering
    \caption{\textbf{Model details.} We report the training details for the models used in this paper in the context of the minimal sample size of \P needed to confidently reject the null hypothesis with \ours.}
    \begin{tabular}{ccccccccc}
    \toprule
    \textbf{} & \textbf{LDM256} & \textbf{U-ViT256} & \textbf{DiT256} & \textbf{U-ViT512} & \textbf{DiT512} & \textbf{U-ViT256-Uncond} & \textbf{U-ViT256-T2I} & \textbf{U-ViT256-T2I-Deep} \\
    \midrule
    \textbf{Model parameters} & 395M & 500M & 675M & 500M & 676M & 44M & 45M & 58M \\
    \textbf{Training steps} & 178k & 500k & 400k & 500k & 400k & 1M & 1M & 1M \\
    \textbf{Batch size} & 1200 & 1024 & 256 & 1024 & 256 & 256 & 256 & 256 \\
    \textbf{Dataset} & ImageNet & ImageNet & ImageNet & ImageNet & ImageNet & COCO & COCO & COCO \\
    \textbf{Dataset size} & 1.2M & 1.2M & 1.2M & 1.2M & 1.2M & 83k & 83k & 83k \\
    \midrule
    \textbf{Min. \P size} & 6000 & 3000 & 2000 & 2000 & 1000 & 300 & 200 & 70 \\

    \bottomrule
    \end{tabular}
    \label{tab:insights}
\end{table*}

\section{Experimental Setup}
\label{app:experimental_setup}
\subsection{Models}
\label{app:models}

\begin{table*}[h!]
\scriptsize
\centering
\caption{\textbf{Feature extraction time for different features.} Time in seconds is given for processing 1 batch of 64 samples on an A100 GPU.}
\label{tab:time}
    \begin{tabular}{ccccccccc}
    \toprule
    \textbf{} & \textbf{LDM256} & \textbf{U-ViT256} & \textbf{DiT256} & \textbf{U-ViT512} & \textbf{DiT512} & \textbf{U-ViT256-Uncond} & \textbf{U-ViT256-T2I} & \textbf{U-ViT256-T2I-Deep} \\
\midrule
\textbf{Denoising Loss} & 4.06 & 10.55 & 12.87 & 9.64 & 40.77 & 2.01  & 2.28  & 2.69 \\
\textbf{SecMI\textsubscript{stat}} & 8.41 & 23.22 & 22.16 & 26.35 & 96.02 & 3.27  & 4.09 & 4.90 \\
\textbf{PIA} & 5.21 & 6.45 & 6.87 & 8.77 & 26.04 & 1.72 & 2.24 & 2.56 \\
\textbf{PIAN} & 5.58 & 6.93 & 7.22 & 9.21 & 28.04 & 1.94 & 2.42 & 2.78 \\
\textbf{Gradient Masking (GM)} & 31.90 &  81.67 &  72.42 & 90.33 & 110.57 & 9.14 & 11.56 & 15.49 \\
\textbf{Multiple Loss (ML)} & 7.11  & 20.28 &  18.42 & 22.55 & 70.28  & 2.92 & 3.45 & 4.26 \\
\textbf{Noise Optim (NO)} & 94.64 & 64.12 & 64.17 & 78.14 & 181.33  & 182.23  & 205.78  & 120.26  \\
\bottomrule
\end{tabular}
\end{table*}

We evaluate the effectiveness of \ours on publicly available state-of-the-art \DMs. This section provides an overview of the models used in our experiments. For more detailed information about these models and their training procedure, readers are encouraged to consult the original papers.
Additionally, we make the model checkpoints readily accessible for download to facilitate the replication of our results. 

All models, with the exception of LDM256 utilize ViT\citep{vit} as the diffusion backbone. LDM256 uses the UNet architecture \cite{unet} instead, being prior work in the area of latent \DMs. 

\textit{LDM256} - a class-conditioned LDM checkpoint provided by \citet{rombach2022high}, trained on ImageNet dataset in 256x256 resolution. The diffuser backbone of this model is a UNet architecture with 395M parameters.

\textit{DiT256}, \textit{DiT512} - class-conditioned DiT-XL/2 checkpoints provided by \citet{peebles2023scalable}, trained on ImageNet dataset in 256x256 and 512x512 resolutions respectively. DiT256 has 675M parameters and DiT512 has 676M parameters.

\textit{U-ViT256}, \textit{U-ViT512}  - class-conditioned U-ViT-Huge/4 checkpoints provided by \citet{bao2023all}, trained on ImageNet dataset in 256x256 and 512x512 resolutions respectively. U-ViT256 has 500.8M parameters and U-ViT512 has 500.9M parameters.

\textit{U-ViT256-T2I} - a text-conditioned U-ViT-Small/4 checkpoint provided by \citet{bao2023all}, trained on COCO dataset in 256x256 resolution.

\textit{U-ViT256-T2I-Deep} - a text-conditioned U-ViT-Small/4-Deep checkpoint provided by \citet{bao2023all}, trained on COCO dataset in 256x256 resolution. The model architecture differs from U-ViT256-T2I by having a larger number of transformer blocks (16 instead of 12). U-ViT256-T2I and U-ViT256-T2I-Deep have 45M and 58M parameters respectively.

\textit{U-ViT256-Uncond} - an unconditioned U-ViT-Small/4 checkpoint trained on COCO dataset in 256x256 resolution. We train the model for 1.000.000 training steps, following the configuration used by \citet{bao2023all} for \textit{U-ViT256-T2I} checkpoint. Namely, we use \textit{AdamW}\citep{adamw} optimizer (lr $2\times10^{-5}$, weight decay 0.03, betas (0.99,0.99)) and batch size 256. U-ViT256-Uncond has 44M parameters.

\subsection{Compute Resources and Feature Extraction Time}
\label{app:compute}
We compute our experiments on A100 80GB NVIDIA GPUs on an internal cluster, amounting to 150 GPU-hours. Training the U-ViT256-Uncond model requires additional 80 GPU-hours. 
We provide the time needed to extract features from a batch of 64 samples in~\cref{tab:time}. Fitting the scoring model $s$ of \ours does not require GPU utilization and can be executed in a negligible time ($<$10 seconds) on a CPU.

\begin{figure*}[t!]
    \centering
    \includegraphics[width=0.8\linewidth]{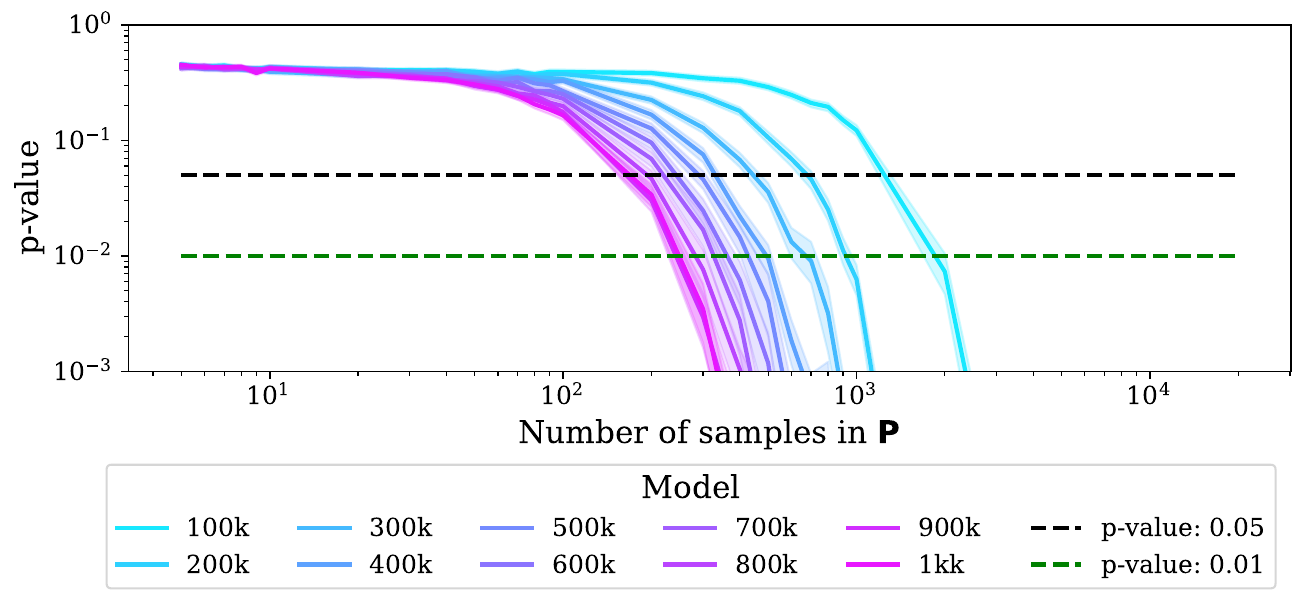}
    \caption{\textbf{Results of \ours for U-ViT256-Uncond for different number of model's training steps} Solid lines indicate aggregated p-values aggregated over 1000 randomized trials for each size of \P, shaded areas around the lines are 95\% confidence intervals. The higher the number of the model's training steps the fewer suspect samples are required from the data owner to confidently reject $H_0$. 
    }
    \label{fig:cdi_results_training_steps}
\end{figure*}

\begin{figure*}[h!]
    \centering
    \includegraphics[width=0.9\linewidth, trim=0cm 0cm 0cm 0cm, clip]{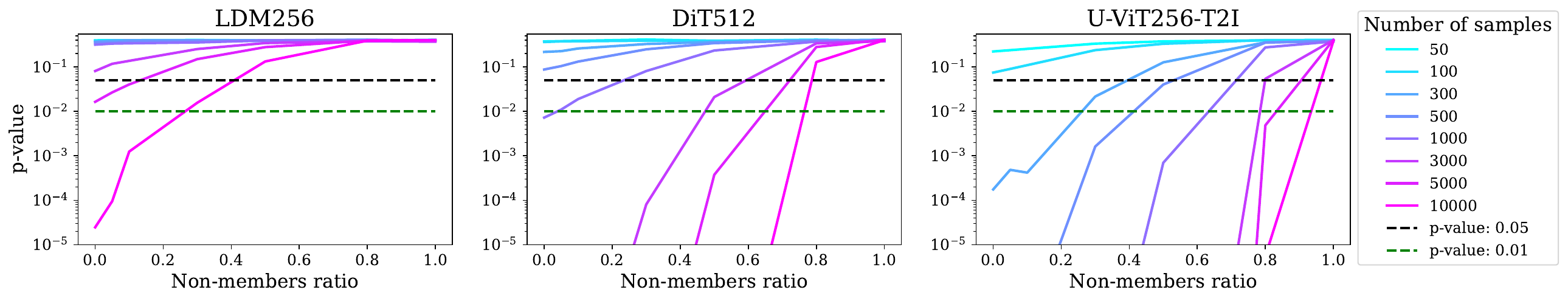}
    
    \caption{\textbf{Impact of non-members ratio in \P on \ours, and resilience against false positives}. The lines represent p-values for a given non-member ratio while varying sizes of \P. 
    Note that for the non-members ratio of 1 (all samples in \P are non-members), the p-values are always significantly above the significance level ($\alpha=0.01$), which means \ours does not return false positive answers. }
    \label{fig:noise_influence_full}
\end{figure*}

\section{Model Details and \ours Effectiveness}
\label{app:insights}
Based on Table~\ref{tab:insights} we make the following observations. (1) For a given DM architecture, trained on a given dataset, the number of samples required for the confident identification of the training data decreases with increasing input resolution. This phenomenon is clearly visible when comparing the required number of samples for U-ViT256 and U-ViT512 and for DiT256 and DiT512 which differ only in the input image resolution (2) Models trained on smaller datasets exhibit stronger signal for identifying training data, as evident when contrasting the results on models trained with ImageNet (LDM, U-ViT256, U-ViT512, DiT256, DiT512) and COCO dataset (U-ViT256-T2I, U-ViT256-T2I-Deep, U-ViT256-Uncond).

\section{Number of Model Training Steps and \ours Effectiveness}
\label{app:ts}

To assess the effect of the number of DM’s training steps on \ours, we conduct the following experiment. We measure the effectiveness of \ours for U-ViT256-Uncond model after every 100,000 training steps. The results in Fig.~\ref{fig:cdi_results_training_steps} confirm that \ours is effective even for models trained for a limited number of steps. With enough samples provided by the data owner $|\P|$ = 2,000, \ours confidently rejects $H_0$ for U-ViT256-Uncond after just 100,000 training steps.

\section{More Results for the Non-members in \P and False Positives}
\label{app:contamination}
We present the additional results on \ours's robustness in cases when \P contains (some) non-member samples in Fig.~\ref{fig:noise_influence_full}. We expand Fig.~\ref{fig:noise_influence} for DiT512 with experimental results for LDM256 and U-ViT256-T2I. 
\ours remains effective even when not all data samples are used as training data. \ie, \P contains a certain ratio of non-members, while the remaining samples in \P are members.
We observe that \ours demonstrates greater robustness in cases when part of \P contains non-members if the data owner supplies larger \P. This is evident in the decreasing p-value at a given non-members ratio as the size of \P increases. Importantly, for the non-members ratio of 1, the p-values are significantly
above the significance level ($\alpha$ = 0.01), which means CDI does not return false positive answers.

\begin{table*}[h!]
\centering
\scriptsize
\caption{\textbf{Performance of CDI under $|\U|$ and $|\P|$ imbalance.} We report the minimal sample size of \P needed to confidently reject the null hypothesis with \ours.}
\begin{tabular}{lrrrrrrr}
\toprule
$|\U|/|\P|$ ratio  & \textbf{LDM256} & \textbf{U-ViT256} & \textbf{DiT256} & \textbf{U-ViT512} & \textbf{DiT512} & \textbf{U-ViT256-T2I} & \textbf{U-ViT256-T2I-Deep} \\
\midrule
1.00 & 6000 & 3000 & 2000 & 2000 & 1000 & 200 & 70 \\
0.90 & 6000 & 3000 & 2000 & 2000 & 2000 & 200 & 70 \\
0.75 & 6000 & 3000 & 2000 & 3000 & 2000 & 200 & 70 \\
0.50 & 6000 & 3000 & 2000 & 3000 & 2000 & 300 & 90 \\
0.25 & 7000 & 4000 & 2000 & 4000 & 2000 & 400 & 200 \\
0.10 & 8000 & 5000 & 4000 & 6000 & 5000 & 800 & 300 \\
\bottomrule
\end{tabular}
\end{table*}

\begin{table*}[h!]
    \centering
    \caption{\textbf{Performance of \ours in gray- vs white-box model access.} 
    We depict the number of samples required from the data owner in \Qsus to reject the null hypothesis for different model access scenarios.
    Our \ours framework remains effective when assuming the more restrictive gray-box model access, provided with larger \P. 
    }
    \scriptsize
    \begin{tabular}{ccccccccc}
    \toprule
    \textbf{Access Type} & \textbf{LDM256} & \textbf{U-ViT256} & \textbf{DiT256} & \textbf{U-ViT512} & \textbf{DiT512} & \textbf{U-ViT256-Uncond} & \textbf{U-ViT256-T2I} & \textbf{U-ViT256-T2I-Deep} \\
    \midrule
    \textbf{Gray-Box Model} & 8000 & 3000 & 2000 & 4000 & 2000 & 400 & 200 & 70 \\
    \textbf{White-Box Model} & 6000 & 3000 & 2000 & 2000 & 1000 & 300 & 200 & 70 \\
    \bottomrule
    \end{tabular}
    \label{tab:graybox}
\end{table*}

\section{Results under $|\P|$ and $|\U|$ imbalance}
In practice, the training-to-test set ratio (\(|U|/|P|\)) is often much smaller than one. We analyze CDI's performance in such scenarios and find it remains effective even when the victim has access to fewer test samples (\(|U|<|P|\)). We report  the number of samples in \(P\) needed to reject \(H_0\) for given ratios (\(|U|=\text{ratio}\times|P|\)). Notably, an extreme set imbalance can be mitigated by increasing \(P\).

\section{MIAs and their Model Access Types}
\label{app:access-type}

We group features, which we use in \ours, into two categories, based on their respective access type.  Because Gradient Masking and Noise Optimization utilize gradient calculations they require white-box access to the \DM's network weights. To execute the remaining feature extraction method \ours needs only gray-box access, \ie the ability to predict the noise added to a clean sample at an arbitrary timestep $t$. We specify the model access type per each feature in \cref{tab:mia_access_type}. 

\begin{table}[h!]
    \scriptsize
    \centering
    \caption{\textbf{Each feature with its Model Access Type.} 
    }
    \begin{tabular}{cc}
    \toprule
    \textbf{MIA} & \textbf{Model Access Type} \\
    \midrule
    Denoising Loss~\citep{carlini2022membership} & gray-box \\
    SecMI$_{stat}$~\citep{duan23bSecMI} & gray-box  \\
    PIA~\citep{kong2024an} & gray-box  \\
    PIAN~\citep{kong2024an} & gray-box  \\ 
    Multiple Loss (ML) & gray-box  \\ \hline
    Gradient Masking (GM) & white-box \\
    Noise Optimization (NO) & white-box \\
    \bottomrule
    \end{tabular}
    \label{tab:mia_access_type}
\end{table}

\section{Gray-box vs. White-box comparison}
\label{app:gray-vs-white}

\ours{} remains effective even in a gray-box model access scenario. We assess the effectiveness of \ours within a gray-box model access scenario, as specified in our threat model (\cref{sec:method}). In this setup, only the original MIA features and Multiple Loss are included, whereas the white-box model access setup leverages the full set of features. Our results in \cref{tab:graybox} show that \ours maintains effectiveness under gray-box access, successfully rejecting the null hypothesis, though it requires a larger sample size in \P.

\section{Analysis of the scoring function}
\label{app:xai_lr}

\begin{figure*}[h!]
    \vspace{0.3cm}
    \centering
    \includegraphics[width=\linewidth]{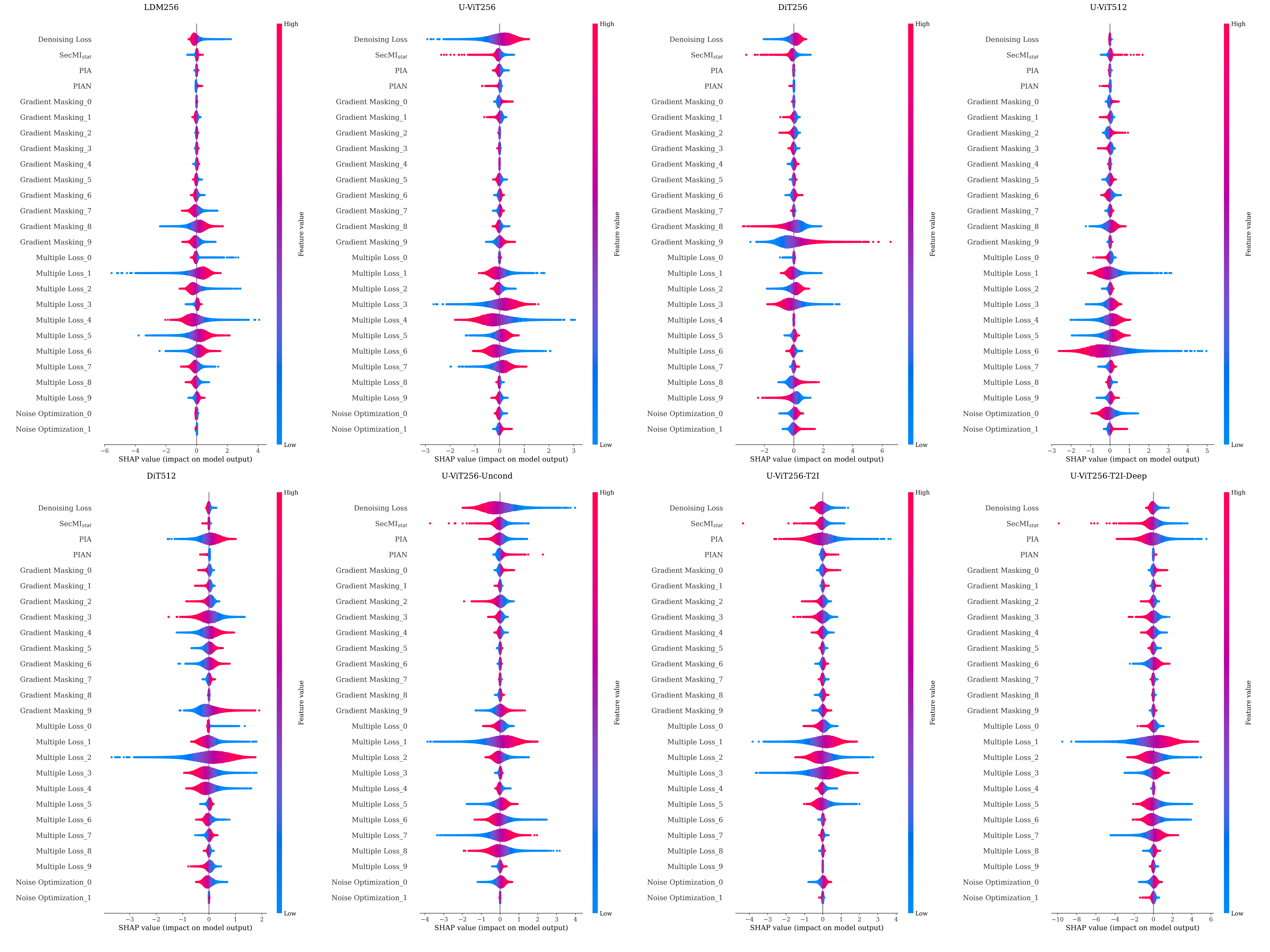}\
    \vspace{-0.5cm}
    \caption{\textbf{SHAPley plots for all models and features.} The scoring functions have been trained using 5,000 members as \Pctrl and 5,000 non-members in \Uctrl, The resulting plots are evaluated on 20,000 members in \Ptest and 20,000 non-members in \Utest to provide the most accurate results.}
    \label{fig:shap}
\end{figure*}

Fig.~\ref{fig:ablation_main} demonstrates the varying impact
of features on CDI performance.
We further aid our analysis of the scoring function by SHAPley~\citep{lundberg2017unified} summary plots in~\cref{fig:shap}.  The results illustrate the model-specific nature of feature
importance. This highlights the necessity of a scoring function that can agnostically learn the optimal feature utilization for each
dataset and DM, leading to more confident and adaptable membership estimations, ultimately enhancing \ours's
effectiveness.

\section{From p-value to \textbf{TPR@FPR=1\%}.}
\label{app:statistical_testing}

The ablation study on the importance of the statistical testing in \ours we conduct in~\cref{sec:stat_cdi_ablation} requires us to transform p-values returned by \ours to \textbf{TPR@FPR=1\%} to directly compare \ours with set-level MIA in~\cref{tab:set_mia_vs_cdi}. We note that the TPR of a statistical test is equivalent to its power. We estimate the power by computing the effect size, using Cohen's d~\cite{cohen1988statistical}, which is defined as $d=\frac{\bar{x_1}-\bar{x_2}}{s}$. $x_1$ and $x_2$ are the observed scores, and $s=\sqrt{\frac{(n_1-1)s_1^2+(n_2-1)s_2^2}{n_1+n_2-2}}$. $n_1$ and $n_2$ are the sizes of the population (here: the number of samples in \P and \U, respectively), and $s_j^2=\frac{1}{n_j-1}\sum_{i=1}^{n_j}(x_{1,j}-\bar{x_j})^2$ for $j=1,2$. After we obtain the effect size, we compute the power of the t-test using a solver, setting $\alpha=0.01$ to get \textbf{TPR@FPR=1\%}.

\section{ROC curves of \ours}
\label{app:cdi_tpr_eval}

\begin{figure*}[h!]
    \centering
    \includegraphics[width=1\linewidth]{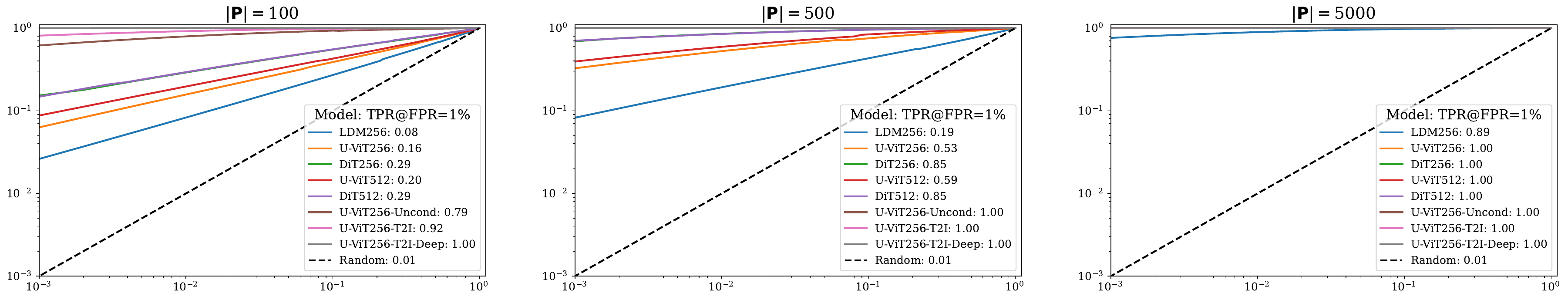}
    \caption{\textbf{ROC curves of \ours.} \ours achieves perfect performance with sufficiently large $|\P|$. Resulting \textbf{TPR@FPR=1\%} align with p-values we report in Fig.~\ref{fig:cdi_results}.}
    \label{fig:di_roc_tpr}
    \vspace{-0.3cm}
\end{figure*}

In~\cref{sec:ours_eval}, we evaluate \ours following the evaluation methodology proposed for DI by \cite{maini2021dataset}. In this section, we additionally extend this evaluation by analyzing \ours through ROC curves and true positive rates (TPR).

We build on~\cref{sec:stat_cdi_ablation} and~\cref{app:statistical_testing}, and obtain TPR at varying FPR by sweeping over $\alpha$ values from 0 to 1. To ensure stability of our results, for each size of \P we sample 1000 subsets from \P and \U, compute the TPR, and finally average the TPR to get the final value.
We visualize our results in~\cref{fig:di_roc_tpr}. Key takeaways from the figure are:
(1) \ours achieves perfect performance given large enough $|\P|$, even at \textbf{FPR=0\%}.
(2) \ours is not over-confident. P-values we report in Fig.~\ref{fig:cdi_results} align with the resulting \textbf{TPR@FPR=1\%}, thanks to the careful choice of $\alpha$ (ref.~\cref{sec:method_stats}). Intuitively, low p-values correspond to high \textbf{TPR@FPR=1\%}. For example, LDM256 for $|$\P$|=5000$ achieves p-values close to $0.01$, and \textbf{TPR@FPR=1\%} of $0.89$.
(3) In effect, \ours is not susceptible to p-value hacking, \ie the improvement in \ours's performance observed in \ours's higher confidence (lower p-value) signifies higher TPR at lower FPR.

\section{MIAs Fail on DI Task}
\label{app:mias_di}

In this experiment, we compare \ours to MIAs on the task of DI. Similarly to set-level MIA we introduce in~\cref{sec:stat_cdi_ablation}, we follow this procedure: We apply a MIA to each sample in the suspect set, returning Positive prediction if the score for any sample exceeds a certain threshold. Repeating this process for every suspect set and varying the threshold yields a ROC curve for each MIA. Finally, we compare these ROC curves against the ROC curve for \ours, obtained as in~App.~\ref{app:cdi_tpr_eval}.

We sample 1000 \P containing only member samples (Positive) and 1000 \P containing only non-member samples (Negative). 
\U remains unchanged, \ie contains only non-members. We vary the size of $P$ for a more thorough analysis and compute ROC curves.

\begin{figure*}
    \centering
    \includegraphics[width=0.9\linewidth,trim=0 5cm 0 0,clip]{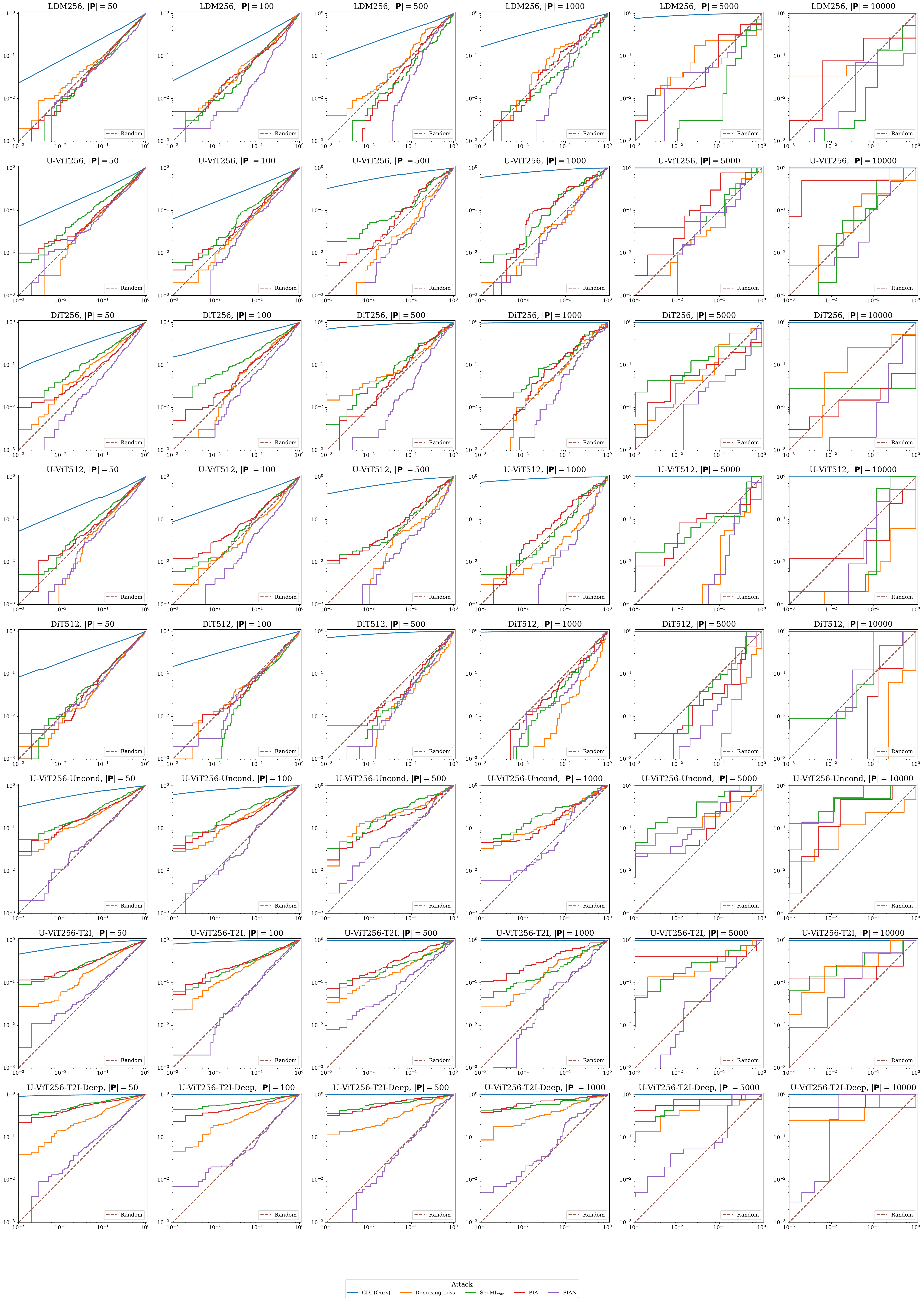}
    \caption{\textbf{Comparison of \ours and MIAs on the DI tasks.}}
    \label{fig:cdi_vs_mias}
\end{figure*}

We demonstrate in Fig.~\ref{fig:cdi_vs_mias} that \ours achieves TPR orders of magnitude higher than MIAs on the DI task.
Applying single-sample MIAs to DI is challenging. Our experiments show MIAs are unstable, with high FPR due to
set-wise confidence swayed by a single high score. Notably, increasing the size of \P does not improve the performance.
\ours's statistical testing is robust by comparing distributions of membership scores and capturing subtle differences. Then, the application of the t-test in \ours 
quantifies these differences, with the p-value serving as a reliable confidence measure.

\section{Further Applicability of Our Novel Features}
\label{app:our_features_analysis}
One of our contributions is the new features introduced in~\cref{sec:our_features}. As we use these features to fit our \textit{scoring function}, we can also use them to perform the default \textit{threshold} MIA. We introduce the detailed results of that experiment in~App.~\ref{app:mias_eval}. Here, we analyze the characteristics of our features.

\subsection{Gradient Masking}
Recall, to obtain this feature we: (1) distort 20\% of the input image's latent based on the absolute gradient values. (2) Use the \DM to reconstruct the distorted part by performing a single denoising step. (3) Compute L2 reconstruction loss over the distorted region.

\begin{figure*}[t!]
    \centering
    \includegraphics[width=1\textwidth]{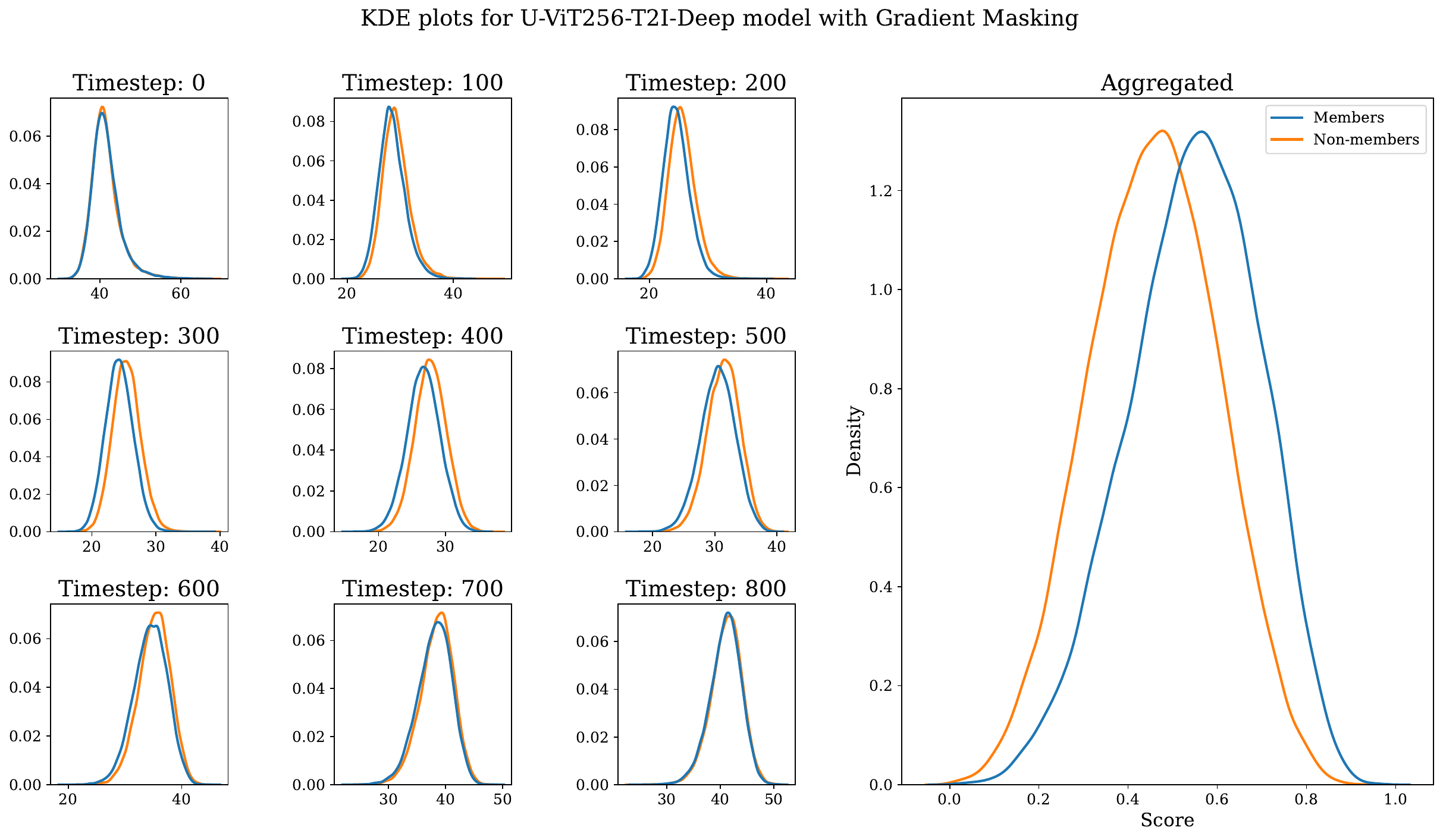}
    \caption{\textbf{KDEplots of the Gradient Masking features for timesteps, and after aggregation.} The model used is U-ViT256-T2I-Deep.}
    \label{fig:gm_kdes}
\end{figure*}

\begin{figure*}[t!]
    \vspace{-1.5cm}
    \centering
    \includegraphics[width=1\textwidth]{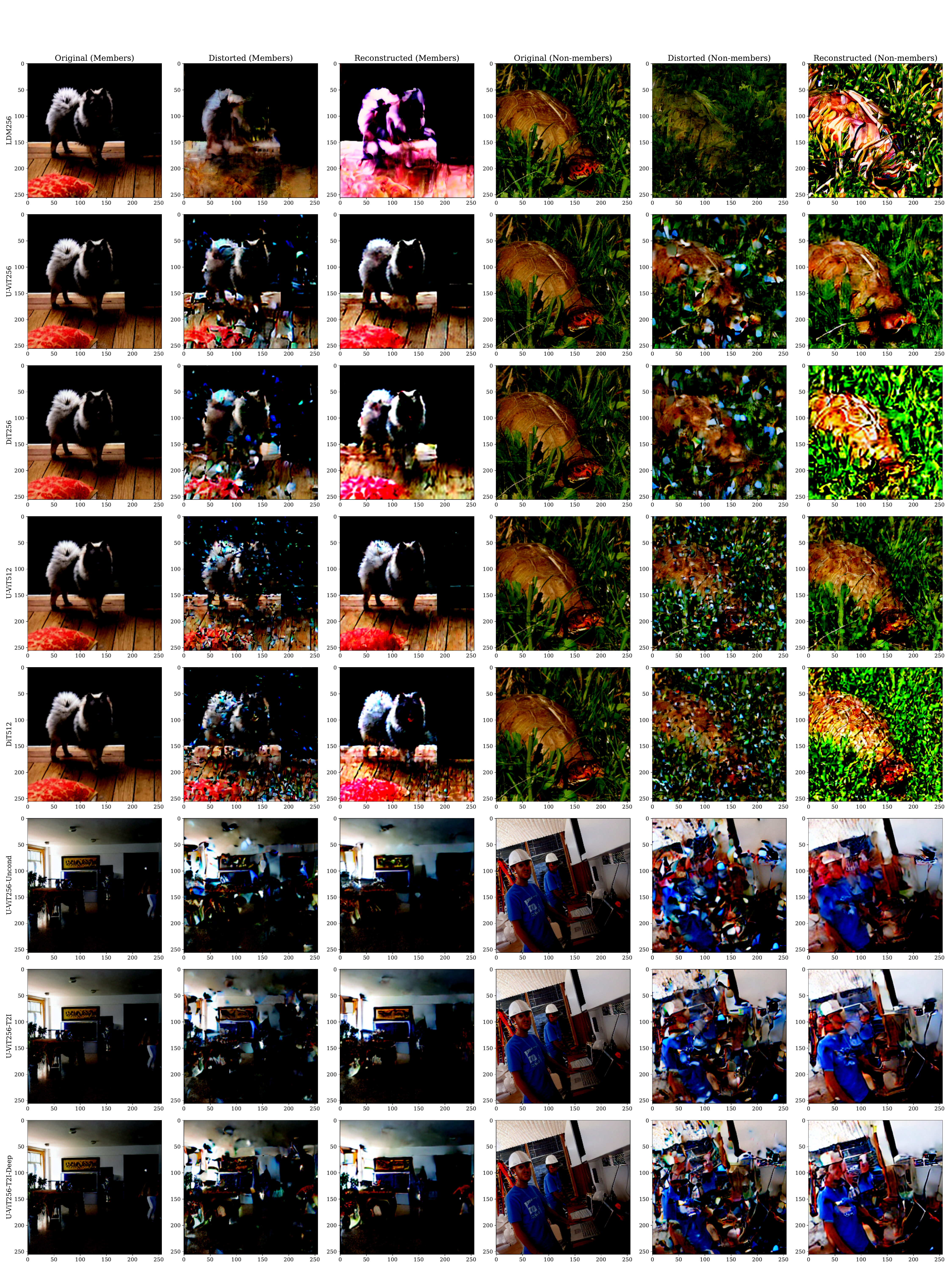}
    \caption{\textbf{Effect of the reconstruction method for the Gradient Masking.} The images in the figure are distorted using noise scale corresponding to the timestep=500, and are denoised using the same timestep. The images differ between models due to the difference in their respective training datasets (ImageNet and COCO2017).}
    \label{fig:gm_images}
\end{figure*}

\paragraph{Visualization.} We visualize the reconstruction effect in~\cref{fig:gm_images}. We observe that: (1) the reconstructed members resemble semantics of the original images better than the reconstructed non-members do. (2) For members and non-members there is a notable decrease in the level of detail in the images after reconstruction, \eg for U-ViT256-T2I-Deep we observe that for the member sample the details of the painting and the table are gone. (3) The crude elements of the reconstructed images stay unchanged, \eg for ImageNet models we can see that the reconstructed image contains a dog (for the member sample), or a turtle in the grass (the non-member). (4) Distortion is not uniform between the models. For the ViT-based models we observe distortions that are spread out throughout the image, while for the LDM (UNet-based model) the distortions are more local.

\begin{figure*}[t!]
    \centering
    \includegraphics[width=1\textwidth]{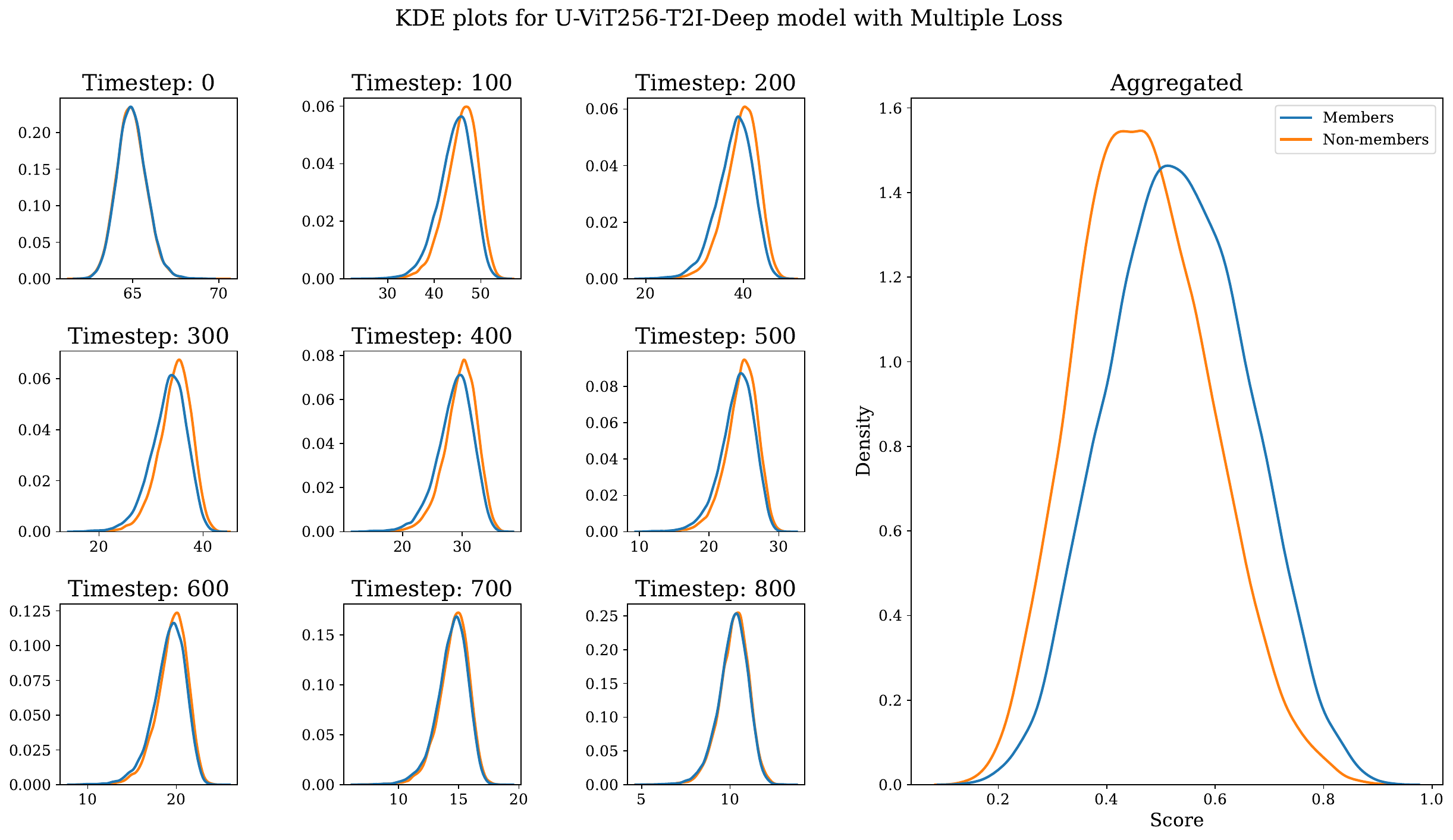}
    \caption{\textbf{KDEplots of the Multiple Loss features for timesteps, and after aggregation.} The model used is U-ViT256-T2I-Deep.}
    \label{fig:ml_kdes}
\end{figure*}

\begin{figure*}[t!]
    \centering
    \includegraphics[width=1\textwidth]{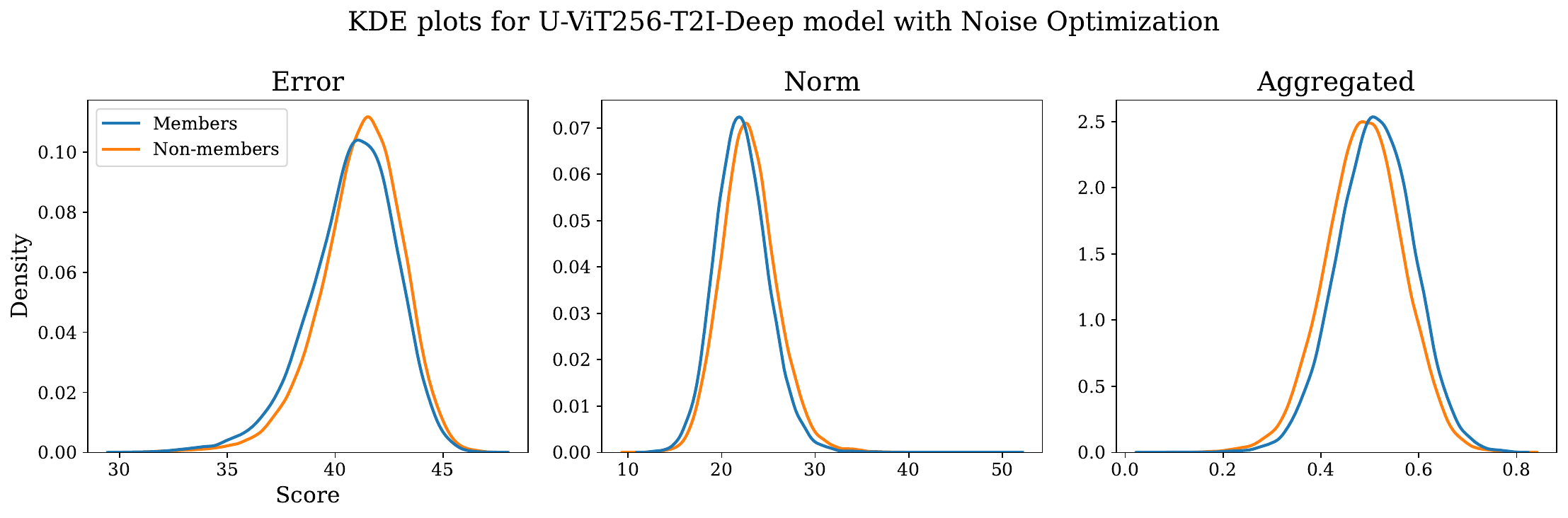}
    \caption{\textbf{KDEplots of the Noise Optim features, and after aggregation.} The model used is U-ViT256-T2I-Deep. Error refers to the reconstruction error obtained after optimization, and Norm is the L2 norm of added perturbation.}
    \label{fig:no_kdes}
\end{figure*}

\paragraph{Features.} In~\cref{fig:gm_kdes} we analyze the distributions of the features (reconstruction errors) throughout various timesteps, and then we compare them with the distribution obtained by aggregating these features with our \textit{scoring function}. We note that the difference between members and non-members is negligible for singular features. However, after we aggregate them we observe a significant difference.

\begin{figure*}[t!]
    \centering
    \includegraphics[width=1\textwidth]{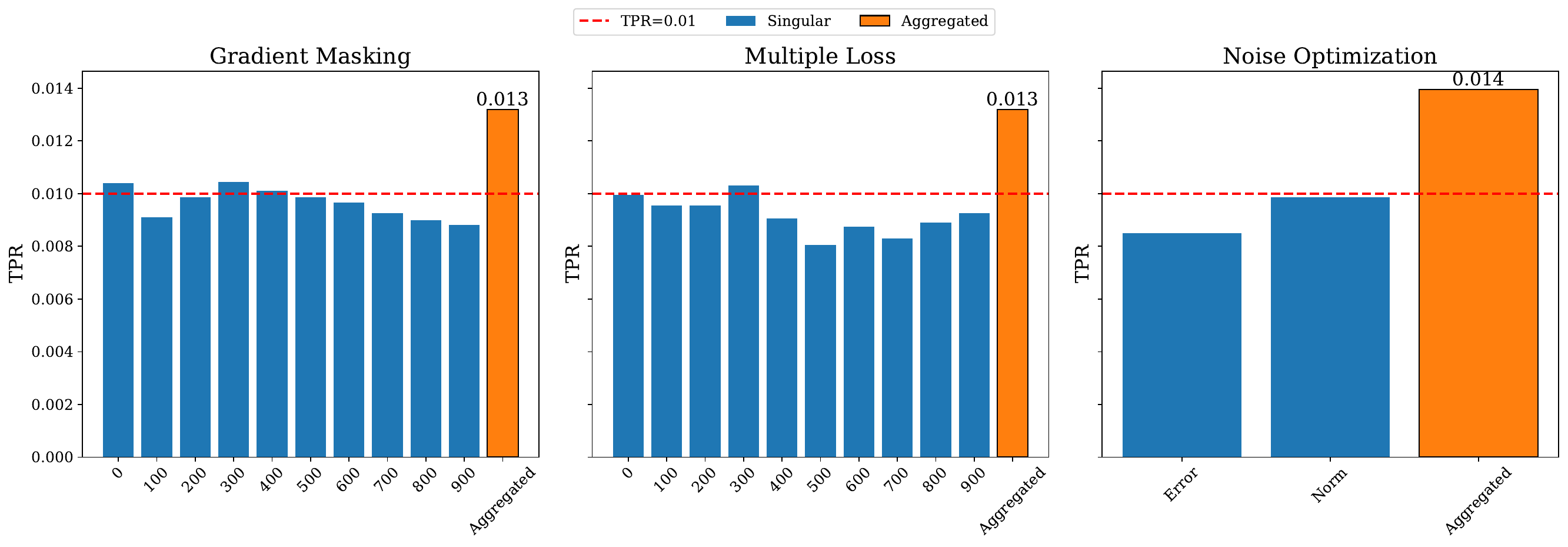}
    \caption{\textbf{TPR@FPR=1\% ($\uparrow$) for singular features, and after aggregation} The model used is LDM256. For Gradient Masking and Multiple Loss the numerical values correspond to the timestep at which the feature is computed.}
    \label{fig:tpr_novel_cmp}
\end{figure*}

\begin{figure*}[t!]
    \centering
    \includegraphics[width=1\textwidth]{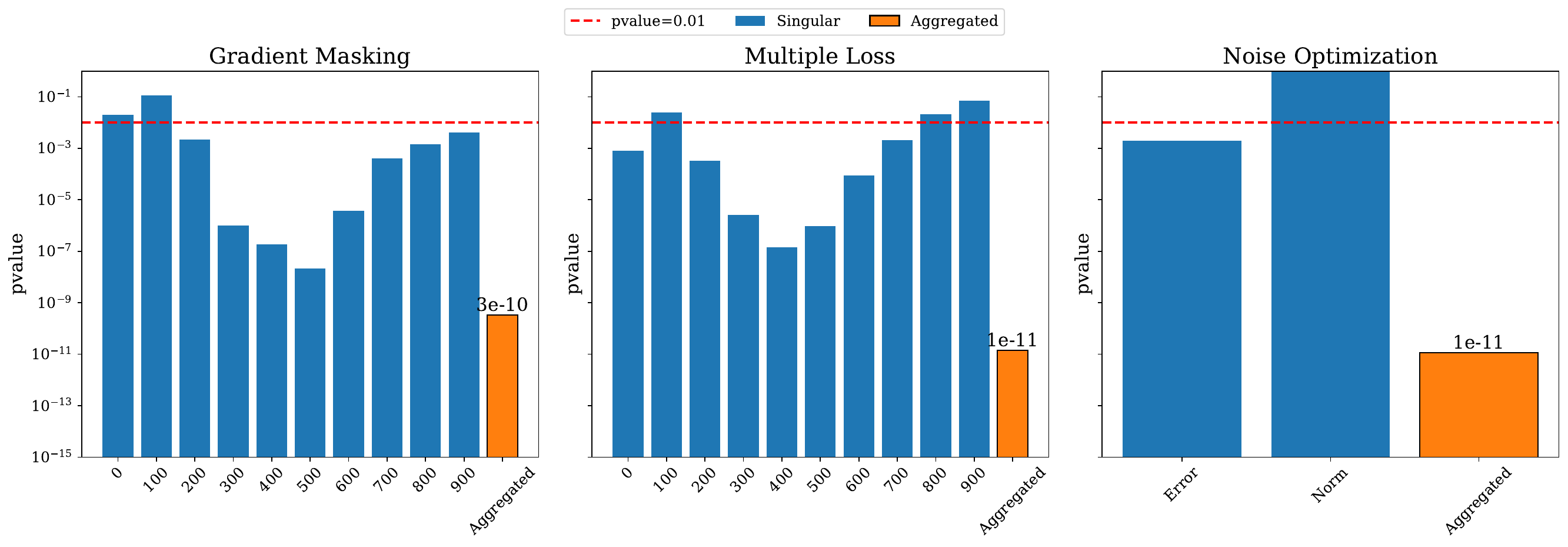}
    \caption{\textbf{P-values ($\downarrow$) for singular features, and after aggregation} The model used is LDM256. For Gradient Masking and Multiple Loss the numerical values correspond to the timestep at which the feature is computed.}
    \label{fig:pvalue_novel_cmp}
\end{figure*}

\subsection{Noise Optim and Multiple Loss}
For these two features our findings are in line with the findings for Gradient Masking. In~\cref{fig:ml_kdes} and~\cref{fig:no_kdes} we visualize the distributions of singular features and the distributions after aggregation.

\subsection{Effect of aggregation on MIA and \ours performance}
We compare the metrics for MIA and \ours when we use a singular feature vs when we aggregate the features. In~\cref{fig:tpr_novel_cmp} we observe that while some singular features barely cross the threshold of random guessing (TPR=1\%), the aggregate provides better results (although the effectiveness is limited). For \ours, in~\cref{fig:pvalue_novel_cmp} we observe that aggregation allows us to lower the p-value by orders of magnitude lower than the best singular feature can.

\section{Further Evaluation of MIAs}
\label{app:mias_eval}

\begin{figure*}[t]
    \centering
    \includegraphics[width=0.95\linewidth]{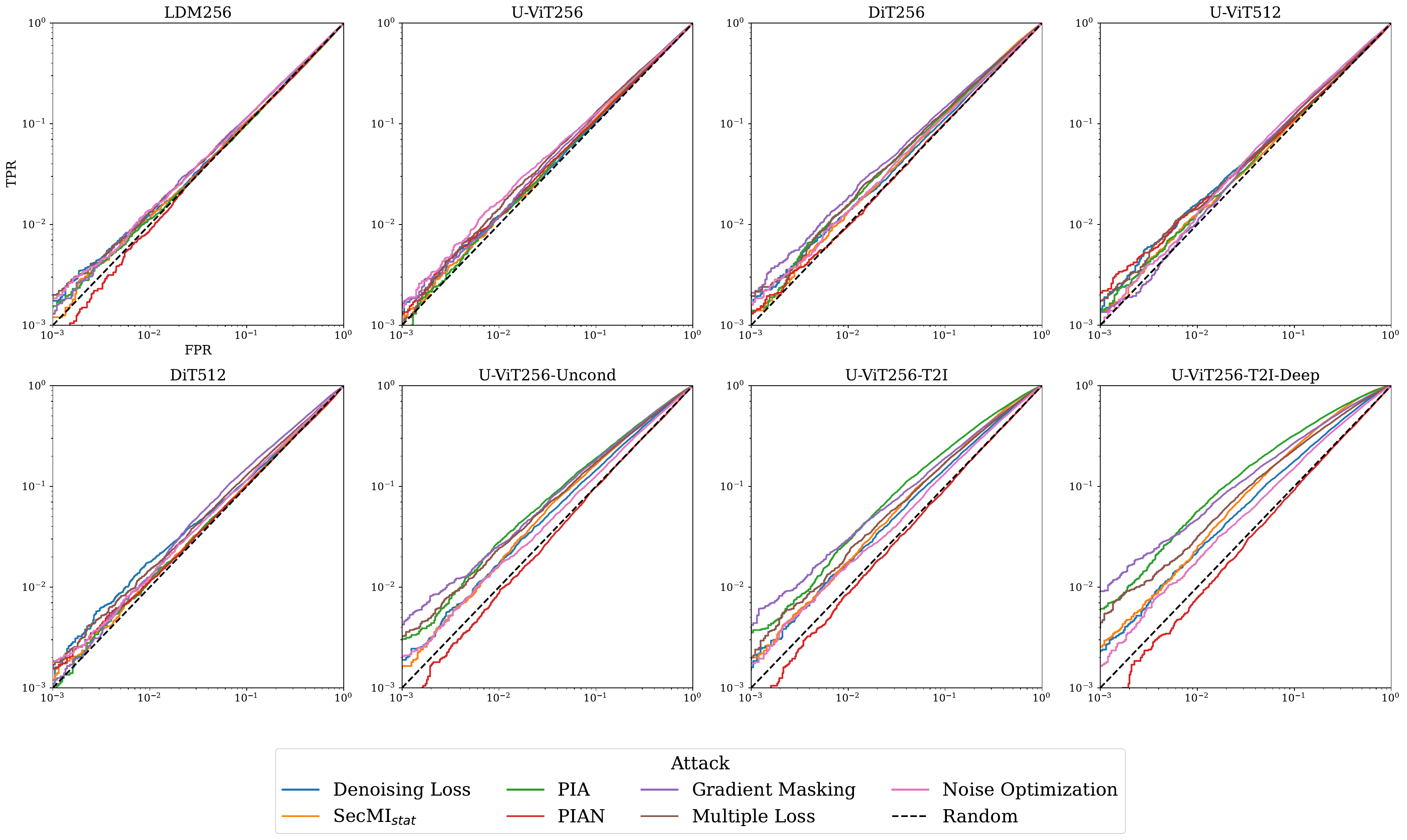}
    \caption{\textbf{ROC curves for MIAs against all models.} X- and Y-axis are in logarithmic scale. We observe that for \DMs trained on ImageNet the TPR in low FPR regime ($<1\%$) is not better than random guessing, while for the models trained on COCO (last three from the right in the bottom row) all methods but PIAN achieve results significantly better than random chance.}
    \label{fig:mia_rocs}
\end{figure*}

The following summarizes an extensive evaluation effort for MIAs utilizing existing, and, for completeness, our proposed novel features used as MIA. We follow identical setting as described in~App.~\ref{app:mias_eval}, and extend the results by Area Under the Curve (AUC) score (Table~\ref{tab:mia_auc_app}), and accuracy (Table~\ref{tab:mia_acc_app}). To perform MIA on novel features we do the following: (1) Fit $s$ on the features extracted from {$|\Pctrl|$} = 5000 and {$|\Uctrl|$} = 5000. (2) Obtain predictions on \Ptest and \Utest. Importantly, $\Ptest\cap \Pctrl=\varnothing$ and $\Utest\cap \Uctrl=\varnothing$. (3) Use these scores to run MIA.
We include accuracy and AUC to better understand the differences between the proposed features, as well as their impact on the \ours. In Fig.~\ref{fig:mia_rocs} we visualize the behavior of the Receiver Operating Characteristic (ROC) curve for all features under MIA setting.

We note that, similarly to Fig.~\ref{fig:ablation_main}, Gradient Masking provides stronger signal than Multiple Loss and Noise Optimization, resulting in higher values of \textbf{TPR@FPR=1\%} (Table~\ref{tab:mia_tpr_app}), AUC, and accuracy in almost all cases. GM underperforms compared to ML only for U-ViT256 and U-ViT512. 

We observe higher performance of MIAs for models trained on smaller datasets, \ie U-ViT256-Uncond, U-ViT256-T2I, U-ViT256-T2I-Deep.

\clearpage

\begin{table*}[h]
\scriptsize
\centering
\caption{\textbf{MIA results at a TPR@FPR=1\%}. Values in the table are in \%.
We include the performance of MIAs using our novel features and note that they perform comparably with the SOTA PIA method, or even outperform it in some cases.
}
\begin{tabular}{ccccccccc}
\toprule
 & \textbf{LDM256} & \textbf{U-ViT256} & \textbf{DiT256} & \textbf{U-ViT512} & \textbf{DiT512} & \textbf{U-ViT256-Uncond} & \textbf{U-ViT256-T2I} & \textbf{U-ViT256-T2I-Deep} \\
\midrule
Denoising Loss~\citep{carlini2022membership} & 1.23{\scriptsize $\pm$0.10} & 1.22{\scriptsize $\pm$0.10} & 1.34{\scriptsize $\pm$0.12} & 1.61{\scriptsize $\pm$0.12} & 1.78{\scriptsize $\pm$0.12} & 1.72{\scriptsize $\pm$0.15} & 1.73{\scriptsize $\pm$0.14} & 2.25{\scriptsize $\pm$0.20} \\
SecMI$_{stat}$~\citep{duan23bSecMI} & 1.17{\scriptsize $\pm$0.11} & 1.17{\scriptsize $\pm$0.11} & 1.31{\scriptsize $\pm$0.14} & 1.26{\scriptsize $\pm$0.10} & 1.16{\scriptsize $\pm$0.13} & 1.67{\scriptsize $\pm$0.19} & 1.78{\scriptsize $\pm$0.14} & 2.41{\scriptsize $\pm$0.24} \\
PIA~\citep{kong2024an} & 1.12{\scriptsize $\pm$0.09} & 1.18{\scriptsize $\pm$0.10} & 1.54{\scriptsize $\pm$0.13} & 1.25{\scriptsize $\pm$0.13} & 1.14{\scriptsize $\pm$0.13} & 2.79{\scriptsize $\pm$0.20} & 2.84{\scriptsize $\pm$0.22} & 5.57{\scriptsize $\pm$0.41} \\
PIAN~\citep{kong2024an} & 0.88{\scriptsize $\pm$0.09} & 1.18{\scriptsize $\pm$0.10} & 0.96{\scriptsize $\pm$0.10} & 1.52{\scriptsize $\pm$0.12} & 1.13{\scriptsize $\pm$0.10} & 0.88{\scriptsize $\pm$0.10} & 0.87{\scriptsize $\pm$0.09} & 0.78{\scriptsize $\pm$0.09} \\
Gradient Masking & 1.32{\scriptsize $\pm$0.12} & 1.20{\scriptsize $\pm$0.12} & 1.83{\scriptsize $\pm$0.15} & 1.07{\scriptsize $\pm$0.12} & 1.28{\scriptsize $\pm$0.12} & 2.56{\scriptsize $\pm$0.16} & 2.98{\scriptsize $\pm$0.22} & 4.71{\scriptsize $\pm$0.30} \\
Multiple Loss & 1.31{\scriptsize $\pm$0.11} & 1.43{\scriptsize $\pm$0.14} & 1.57{\scriptsize $\pm$0.14} & 1.43{\scriptsize $\pm$0.10} & 1.48{\scriptsize $\pm$0.13} & 2.39{\scriptsize $\pm$0.18} & 2.19{\scriptsize $\pm$0.26} & 3.10{\scriptsize $\pm$0.28} \\
Noise Optimization & 1.39{\scriptsize $\pm$0.11} & 1.67{\scriptsize $\pm$0.13} & 1.35{\scriptsize $\pm$0.12} & 1.25{\scriptsize $\pm$0.16} & 1.25{\scriptsize $\pm$0.13} & 1.63{\scriptsize $\pm$0.13} & 1.66{\scriptsize $\pm$0.15} & 1.81{\scriptsize $\pm$0.17} \\
\bottomrule
\end{tabular}
\label{tab:mia_tpr_app}
\end{table*}

\begin{table*}[h]
\scriptsize
\centering
\caption{\textbf{Accuracy of MIA on all features}. Values are in \%. 
Here we observe that all novel features outperform already existing ones. Note that for all models trained on ImageNet (first five columns from the left) we observe results very close to 50\%, essentially random guessing. For models trained on COCO (remaining three columns on the right) we observe an improvement for Gradient Masking and Multiple Loss, while the MIAs from~\cref{sec:mia_features} remain close to 50\%. 
}
\label{tab:mia_acc_app}
\begin{tabular}{ccccccccc}
\toprule
 & \textbf{LDM256} & \textbf{U-ViT256} & \textbf{DiT256} & \textbf{U-ViT512} & \textbf{DiT512} & \textbf{U-ViT256-Uncond} & \textbf{U-ViT256-T2I} & \textbf{U-ViT256-T2I-Deep} \\
\midrule
Denoising Loss~\citep{carlini2022membership} & 50.02{\scriptsize $\pm$0.01} & 50.07{\scriptsize $\pm$0.03} & 50.09{\scriptsize $\pm$0.03} & 50.14{\scriptsize $\pm$0.04} & 50.20{\scriptsize $\pm$0.04} & 50.01{\scriptsize $\pm$0.01} & 50.01{\scriptsize $\pm$0.01} & 50.01{\scriptsize $\pm$0.01} \\
SecMI$_{stat}$~\citep{duan23bSecMI} & 50.03{\scriptsize $\pm$0.03} & 50.04{\scriptsize $\pm$0.04} & 50.17{\scriptsize $\pm$0.20} & 50.03{\scriptsize $\pm$0.09} & 49.97{\scriptsize $\pm$0.04} & 52.92{\scriptsize $\pm$2.89} & 50.40{\scriptsize $\pm$0.50} & 53.57{\scriptsize $\pm$3.55} \\
PIA~\citep{kong2024an} & 50.01{\scriptsize $\pm$0.01} & 50.02{\scriptsize $\pm$0.03} & 50.28{\scriptsize $\pm$0.06} & 50.02{\scriptsize $\pm$0.03} & 50.03{\scriptsize $\pm$0.03} & 50.07{\scriptsize $\pm$0.02} & 50.03{\scriptsize $\pm$0.01} & 50.03{\scriptsize $\pm$0.01} \\
PIAN~\citep{kong2024an} & 49.51{\scriptsize $\pm$0.12} & 49.87{\scriptsize $\pm$0.06} & 49.81{\scriptsize $\pm$0.07} & 49.82{\scriptsize $\pm$0.17} & 49.80{\scriptsize $\pm$0.13} & 50.02{\scriptsize $\pm$0.15} & 50.10{\scriptsize $\pm$0.15} & 50.05{\scriptsize $\pm$0.12} \\
Gradient Masking & 50.00{\scriptsize $\pm$0.01} & 50.01{\scriptsize $\pm$0.01} & 51.61{\scriptsize $\pm$0.46} & 50.00{\scriptsize $\pm$0.01} & 50.78{\scriptsize $\pm$0.13} & 53.60{\scriptsize $\pm$0.60} & 55.26{\scriptsize $\pm$0.68} & 61.85{\scriptsize $\pm$0.30} \\
Multiple Loss & 50.07{\scriptsize $\pm$0.07} & 50.00{\scriptsize $\pm$0.01} & 50.08{\scriptsize $\pm$0.06} & 50.00{\scriptsize $\pm$0.00} & 50.06{\scriptsize $\pm$0.06} & 52.75{\scriptsize $\pm$0.19} & 53.70{\scriptsize $\pm$0.29} & 59.79{\scriptsize $\pm$0.26} \\
Noise Optimization & 50.00{\scriptsize $\pm$0.00} & 50.00{\scriptsize $\pm$0.00} & 50.00{\scriptsize $\pm$0.00} & 50.09{\scriptsize $\pm$0.04} & 50.00{\scriptsize $\pm$0.00} & 50.03{\scriptsize $\pm$0.02} & 50.24{\scriptsize $\pm$0.06} & 52.72{\scriptsize $\pm$0.20} \\
\bottomrule
\end{tabular}
\end{table*}

\begin{table*}[h]
\scriptsize
\centering
\caption{\textbf{AUC score for MIAs}. 
We observe that for this metric PIA outperforms all standalone features for models trained on COCO  (last three columns on the right) while MIAs based on Gradient Masking, Multiple Loss and Noise Optim achieve better performance for models trained on ImageNet (first five columns on the left) in almost all cases.
}
\label{tab:mia_auc_app}
\begin{tabular}{ccccccccc}
\toprule
 & \textbf{LDM256} & \textbf{U-ViT256} & \textbf{DiT256} & \textbf{U-ViT512} & \textbf{DiT512} & \textbf{U-ViT256-Uncond} & \textbf{U-ViT256-T2I} & \textbf{U-ViT256-T2I-Deep} \\
\midrule
Denoising Loss~\citep{carlini2022membership} & 50.55{\scriptsize $\pm$0.28} & 50.29{\scriptsize $\pm$0.30} & 51.71{\scriptsize $\pm$0.29} & 49.99{\scriptsize $\pm$0.29} & 50.19{\scriptsize $\pm$0.29} & 56.47{\scriptsize $\pm$0.28} & 57.38{\scriptsize $\pm$0.28} & 60.77{\scriptsize $\pm$0.29} \\
SecMI$_{stat}$~\citep{duan23bSecMI} & 49.59{\scriptsize $\pm$0.28} & 53.06{\scriptsize $\pm$0.29} & 55.22{\scriptsize $\pm$0.28} & 50.92{\scriptsize $\pm$0.30} & 50.69{\scriptsize $\pm$0.30} & 59.28{\scriptsize $\pm$0.29} & 61.56{\scriptsize $\pm$0.28} & 69.20{\scriptsize $\pm$0.26} \\
PIA~\citep{kong2024an} & 49.02{\scriptsize $\pm$0.29} & 51.65{\scriptsize $\pm$0.29} & 53.07{\scriptsize $\pm$0.29} & 50.79{\scriptsize $\pm$0.29} & 49.98{\scriptsize $\pm$0.29} & 59.97{\scriptsize $\pm$0.27} & 63.99{\scriptsize $\pm$0.28} & 71.18{\scriptsize $\pm$0.25} \\
PIAN~\citep{kong2024an} & 49.41{\scriptsize $\pm$0.28} & 50.69{\scriptsize $\pm$0.29} & 49.88{\scriptsize $\pm$0.28} & 49.82{\scriptsize $\pm$0.28} & 49.07{\scriptsize $\pm$0.29} & 49.63{\scriptsize $\pm$0.30} & 49.68{\scriptsize $\pm$0.29} & 49.35{\scriptsize $\pm$0.28} \\
Gradient Masking & 51.66{\scriptsize $\pm$0.29} & 53.06{\scriptsize $\pm$0.29} & 54.83{\scriptsize $\pm$0.29} & 53.67{\scriptsize $\pm$0.29} & 56.05{\scriptsize $\pm$0.28} & 59.36{\scriptsize $\pm$0.29} & 59.88{\scriptsize $\pm$0.28} & 66.80{\scriptsize $\pm$0.27} \\
Multiple Loss & 51.77{\scriptsize $\pm$0.30} & 53.63{\scriptsize $\pm$0.29} & 54.16{\scriptsize $\pm$0.30} & 52.30{\scriptsize $\pm$0.30} & 52.98{\scriptsize $\pm$0.29} & 58.17{\scriptsize $\pm$0.29} & 58.93{\scriptsize $\pm$0.28} & 64.26{\scriptsize $\pm$0.27} \\
Noise Optimization & 51.85{\scriptsize $\pm$0.29} & 52.73{\scriptsize $\pm$0.29} & 52.15{\scriptsize $\pm$0.29} & 54.09{\scriptsize $\pm$0.29} & 51.93{\scriptsize $\pm$0.29} & 54.28{\scriptsize $\pm$0.28} & 55.51{\scriptsize $\pm$0.29} & 58.04{\scriptsize $\pm$0.27} \\
\bottomrule
\end{tabular}
\end{table*}

\end{document}